\documentclass{elsarticle}

\usepackage{rotating}
\usepackage{lineno,hyperref}

\usepackage{pgfplotstable}

\usepackage{amsthm,amsmath}
\usepackage{amssymb}
\usepackage[utf8]{inputenc} 
\usepackage[]{algorithm,algpseudocode}
\usepackage{tikz}
\usetikzlibrary{positioning}
\usepackage{booktabs}
\usepackage{pgfplots}
\pgfplotsset{compat=1.7}

\usepackage{tablefootnote}

\usepackage{collcell}
\usepackage{hhline}
\usepackage{pgf}
\usepackage{multirow}
\usepackage{colortbl}
\usepackage[caption=true]{subfig}

\tikzset{font={\fontsize{8pt}{12}\selectfont}}

\usepackage{graphicx}
\graphicspath{ {images/} }

\DeclareMathOperator*{\argmin}{arg\,min}

\def\colorModel{rgb} 

\newcommand\ColCell[1]{
  \pgfmathparse{#1<50?1:0}  
    \ifnum\pgfmathresult=0\relax\color{white}\fi
  \pgfmathsetmacro\compA{#1/100}      
  \pgfmathsetmacro\compB{1-#1/100}      
  \pgfmathsetmacro\compC{1}      
  \edef\x{\noexpand\centering\noexpand\cellcolor[\colorModel]{\compA,\compB,\compC}}\x #1
  } 
\newcolumntype{E}{>{\collectcell\ColCell}m{0.4cm}<{\endcollectcell}}  

\pgfplotstableset{
    color cells/.style={
        col sep=comma,
        string type,
        postproc cell content/.code={%
                \pgfkeysalso{@cell content=\rule{0cm}{2.4ex}\cellcolor{blue!##1}\pgfmathtruncatemacro\number{##1}\ifnum\number>50\color{white}\fi##1}%
                },
        columns/x/.style={
            column name={},
            postproc cell content/.code={}
        }
    }
}

\begin{document}

\begin{frontmatter}

\title{Disparity-Augmented Trajectories \\for Human Activity Recognition}


\author[uwinadd]{Pejman Habashi\fnref{myfootnote}}
\ead{habashi@uwindsor.ca}
\fntext[myfootnote]{Correspoing Author}

\author[uwinadd]{Boubakeur Boufama}
\ead{boufama@uwindsor.ca}

\author[uwinadd]{Imran Shafiq Ahmad}
\ead{imran@uwindsor.ca}

\address[uwinadd]{School Of Comupter Science, University of Windsor, 401 Sunset Ave, Windsor, ON, CANADA N9B 3P4}

\begin{abstract}
	Numerous methods for human activity recognition have been proposed in the past two decades. Many of these methods are based on sparse representation, which describes the whole video content by a set of local features. Trajectories, being mid-level sparse features, are capable of describing the motion of an interest-point in 2D space. 2D trajectories might be affected by viewpoint changes, potentially decreasing their accuracy. In this paper, we initially propose and compare different 2D trajectory-based algorithms for human activity recognition. Moreover, we propose a new way of fusing disparity information with 2D trajectory information, without the calculation of 3D reconstruction. The obtained results show a 2.76\% improvement when using disparity-augmented trajectories, compared to using the classical 2D trajectory information only. Furthermore, we have also tested our method on the challenging Hollywood 3D dataset, and we have obtained competitive results, at a faster speed. 
\end{abstract}

\begin{keyword}
\texttt{Human Activity Recognition \sep Disparity-Augmented Trajectory \sep Video Rectification \sep Video Content Analysis}
\end{keyword}

\end{frontmatter}


\section{Introduction}
\label{sec::Introduction}
Automatic human activity recognition (HAR) is the process of automatically labeling the videos containing human movements with the corresponding action names. Johansson et al. \cite{johansson1973visual} carried out an interesting experiment where, they attached markers to human joints before recording the movement of these markers in a video. In almost all cases, human subjects could say that the tags were attached to a human body. Furthermore, they were able to correctly guess the type of activity the actor was doing. Although this experiment  clearly proves that the task of HAR is rather trivial for humans, it does not tell us whether the human brain uses 2D trajectories of these markers or creates a 3D trajectory model, before the actual recognition process.

Numerous approaches have been proposed for solving the problem of human activity recognition. The use of sparse representation, where each video is represented by a set of independent features, has gained a lot of popularity. Most of the features proposed in the literature focused on low-level features~\cite{laptev2003interest,laptev2005space,dollar2005behavior,
laptev2008learning,bregonzio2009recognising,wang2009evaluation,pervs2010histograms}, that are directly extracted from the pixel information. Other works have used higher level features, such as main joint locations~\cite{li2010object,sadanand2012action,yao2011does,jhuang2013towards}, with the assumption that high-level features would yield better results, if they were accurately known.
Unfortunately, extracting high-level features from cluttered scenes is a non-trivial task. It might require the use of specialized equipments, like color markers on human joints~\cite{ofli2013berkeley} for example, or Microsoft Kinect sensor or other active sensors, to create a depth map and to extract human skeleton from it~\cite{shotton2013real, barnachon2014ongoing}.
On the other hand, it is relatively easy to build a trajectory by tracking a set of 2D interest points across video frames.
When compared to other sparse representation methods, trajectories are mid-level features that can yield competitive results. Recently, \cite{habashi2017trajectory} proposed and compared different 2D trajectory-based HAR algorithms. In a separate work, \cite{habashi2017better} proposed a better trajectory shape descriptor to be used for HAR. 

We have used disparity as another new feature to boost the performance of trajectory-based methods. 
To calculate disparities, two slightly different views of the subject are required. First, 2D trajectories are extracted from the left and right videos. Then, by matching these trajectories and mapping them to the rectified image planes, a disparity-augmented trajectory is created.

This paper demonstrates that adding the disparity information to the 2D trajectories, can be beneficial for human activity recognition. In particular, disparity-augmented trajectories have improved classification rates by 2.76\% in our tests.  

Both 2D and disparity-augmented trajectories are made of pixel locations across frames. To be used for classification, a descriptor, that can discriminate between different trajectories, should be defined based on the shape of trajectories. The descriptors used in this work were inspired by the ones used in \cite{habashi2017better}.

We have also improved the performance of our proposed method by limiting the 
processing to the regions of interest, instead of the whole images. Our regions
of interest consist of the parts in the video frames that contain movement.
In particular, the graph connected component analysis algorithm was used to select the active areas in frames.

It is also worth noting that the use of trajectories proposed in this paper, is independent from the HAR task, and it can be applied to any other video categorization problem. However, to show the effectiveness of our proposed algorithms and compare them with the state of the art methods, we have applied and tested them on human activity recognition application. 


In summary, this paper has the following contributions:

\begin{itemize}
	\item A new method to extract disparity-augmented trajectories from stereo videos
	\item Extension of the trajectory shape descriptor to higher dimensions
	\item A method to rectify stereo videos
	\item A new dataset of stereo videos 
	\item Comparison of 2D trajectories versus disparity-augmented trajectories in the HAR domain
\end{itemize}









The remainder of this paper is organized as follows. Section~\ref{sec::backgroundWorks} reviews related research works and Section~\ref{sec::preprocessing} describes the method used for detecting human activity areas in video frames. Section~\ref{sec::trajectoriesForHumanActivityRecognition} provides details for three different algorithms we have used to extract 2D trajectories and describes how the disparity information is added to the 2D trajectories. Section~\ref{sec::trajectoryShapeEncoding} describes the proposed trajectory shape encoding algorithm. Details of the learning method used can be found in section~\ref{sec::learningPhase}. Section~\ref{sec::experiments} and Section~\ref{sec::conclusion} present the experimental results and the conclusion, respectively.


\section{Background works}
\label{sec::backgroundWorks}
Trajectories have proven to be useful for aligning consecutive frames before extracting low-level features~\cite{wang2011action}. Even the extraction of deep learning feature vectors benefited from trajectory alignment~\cite{wang2015action}. Trajectory shapes can also be used directly for human activity recognition. 

Wang et al.~\cite{wang2013dense,wang2013action,wang2011action} exploited trajectories in separate contributions. In their works, a grid was used to dense sample video frames. Eigenvalues of the autocorrelation matrix was utilized to filter out the samples that were not easy to track. Dense optical flow field, proposed by Farnback~\cite{farneback2003two}, was applied to track these sample points in time. This flow was then employed to align the interest points neighborhoods before calculating the HOG and HOF features. They also proposed another trajectory shape descriptor, that did not outperform the other two.

Mademlis et al.~\cite{mademlis2016exploiting} used disparity information to calculate HOG, HOF and MBH in different disparity zones before encoding them for activity recognition. Although their method improves the performance, but their use of disparity is limited to few disparity zones. Arguably the disparity can be used more efficiently for encoding task.


Hadfield et al.~\cite{hadfield2017hollywood} used 3D Hollywood movies to create a challenging stereo dataset for human activity recognition. The authors estimated the calibration information using RANSAC method and repeating the process 100 times, before selecting the best estimation. Then, the extracted 3D information extracted was used to calculate 3.5D interest points. They have defined a 3D motion descriptors for each of these feature points and, they have normalized it to remove the effect of different camera rotations.

Matikainen et al.~\cite{matikainen2009trajectons} used the technique of Kanade Lucas Tomasi (KLT)~\cite{lucas1981iterative} 
to track a number of points and, created a trajectory for each of these points. Then, they used K-Means method to cluster the obtained trajectories in different clusters (words). They have also proposed to augment these trajectories by adding some affine transformation information, which represents the motion of various parts of the body. Finally, they have used a standard bag of words (BOW) method and SVM for clustering.

In another similar work, Messing et al.~\cite{messing2009activity} used KLT to track keypoints of a video and created a generative model on the velocity history of these keypoints.

Sun et al.~\cite{sun2009hierarchical} proposed to track Scale Invariant Feature Transform (SIFT) points. They have used SIFT descriptors to match each keypoints across the frames. They have extracted features at different levels and used multichannel nonlinear SVM for human activity recognition.

More recently,~\cite{habashi2017trajectory,habashi2017better} demonstrated that good sparse trajectories could produce competitive results to low-level features, but with less computations. Besides, trajectories are a better choice for HAR as they encode the motion of a body, while low-level features usually encode the texture or movement within small neighborhoods in spatiotemporal space. This makes low-level features more dataset dependent.






\section{Preprocessing}
\label{sec::preprocessing}
In order to reduce the overall processing time, we have developed a simple, yet effective, method for detecting the regions of interest (moving parts in the videos). The steps below describe how we detect and remove static (stable) regions from videos.

\begin{enumerate}
	\item Estimate background with a mixture of Gaussian
	\item Subtract estimated background from current frame
	\item Highlight the moving parts of video by erosion and dilation operations
	\item Extract the contours of motion 
	\item Find rectangular regions of interest as follows:
	\begin{enumerate}
		\item Find a bounding box for each contour
		\item Create a graph, where each node represents a bounding box and, if two boxes overlap or are close enough, have an edge between them

		\item Use connected component labeling algorithm similar to~\cite{vincent1991watersheds} to find the connected components of this graph 
		\item Combine the boxes of each connected components. Each combined box represents a separate region of interest
	\end{enumerate}
\end{enumerate}


The above algorithm allows the extraction of all non-static (motion) areas.

\section{Trajectories for Human Activity Recognition}
\label{sec::trajectoriesForHumanActivityRecognition}

Trajectories are defined as the trail of 2D or 3D spatial feature points in time. The disparity-augmented trajectories are similar to 3D trajectories except that they have the disparity in addition to 2D information. Formally, a trajectory is defined as an ordered list of locations, sampled over $l+1$ time steps, where $l$ is the length of the trajectory. In a single video, the frame rate determines the distance between sampling times, so a trajectory $T$ in dimension $n$ can be defined as:

\begin{equation}
	\label{equ::Trajectory}
	T = (p_0, p_1, p_2, ..., p_l), p_i \in \mathbb{R}^{n}, i=0 ... l
\end{equation}

Note that throughout this paper we assume that $n\in{\mathbb{N}}$, but in practice 
and in our tests, $n\in{\{2,3\}}$.

To create a disparity-augmented trajectory, the corresponding 2D trajectories from two views of a subject are extracted and combined. Section~\ref{sec::2dTrajectoryExtraction} provides details of how these 2D trajectories are extracted. Then, section~\ref{sec::3dTrajectoryExtraction} explains how disparity is added to our 2D trajectories.

\subsection{2D Trajectory Extraction}
\label{sec::2dTrajectoryExtraction}
A 2D trajectory $T$ is an ordered list of 2D spatial coordinates $p=(x,y)$ in $l+1$ consecutive frames, formally defined as:

\begin{equation}
  \label{equ::Trajectory2D}
  T=(p_0, p_1, p_2, ..., p_l), p_i \in \mathbb{R}^{2}, i=0..l
\end{equation}

Authors in \cite{habashi2017trajectory} compared three different trajectory extraction algorithms and showed that a combination of FAST corner detector and Farnback optical flow for trajectory extraction outperformed other trajectory extraction algorithms. More details are given below for each of the three methods.

\subsubsection{Interest Point Tracking}
\label{sub::interestPointTracking}
We refer to this method as \textit{``Interest Point Tracking''} (IP).
Starting from the first frame, interest points are extracted then tracked across frames to make trajectories.
When a trajectory reaches a length of $l+1$, it is considered a complete trajectory.

$T$ and $\Psi$ are two sets to keep trajectories and matched feature points, respectively. 
For each frame $I_t$, the set of feature points $P_t$ and the set of next frame feature points $P_{t+1}$ are extracted.

For each member of $P_t$, a feature descriptor is calculated based on the appearance of the interest point neighborhood. Again, the algorithm could consider any feature descriptor.

We define a mapping $\Psi_t: P_t \to \mathbb{R}^k$, where

\begin{equation}
  \label{equ::FeatureDescriptors}
  \Psi_t = \{(p, v)| p \in P_t, v \in \mathbb{R}^k \}
\end{equation}
$t$ is the frame number, $v$ is the feature descriptor vector for point $p$ and $k$ is the dimension of the descriptor (e.g., for standard SIFT descriptor $k=128$).

The neighborhood of a point $p_i$ is given by
\begin{equation}
  N_{t}(p_0)=\{p~|~p \in P_{t},~\Delta_1(p, p_0) \leq \lambda_1\}
\end{equation}
where $\Delta_1(.)$ is a distance measure and $\lambda_1$ determines the radius of neighborhood. 
In our implementation, we have used the Manhattan distance. 

The best match for $p_0$ is found within the neighborhood $N_{t}(p_0)$, based on
the appearance of descriptors.
To do so, for each $p_i \in P_{t-1}$, a mapping $M_{t}(p_0): N_{t}(p_0) \to \mathbb{R}$ is defined as follow:

\begin{equation}
\centering
  M_{t}(p_0)=\{(p, d)| p \in N_{t}(p_0),
  d = \Delta_2(\Psi_{t-1}(p), \Psi_{t}(p_0))\}
\end{equation}
where $\Delta_2(.)$ is a distance measure in descriptors space.

In our case, we used the Euclidean distance, where the closest match is considered the best match.

\begin{equation}
  BestMatch = \argmin\{  M_{t+1}(p_i)\}
\end{equation}

\begin{figure*}[!htbp]
	\centering
	\includegraphics[width=0.325\linewidth]{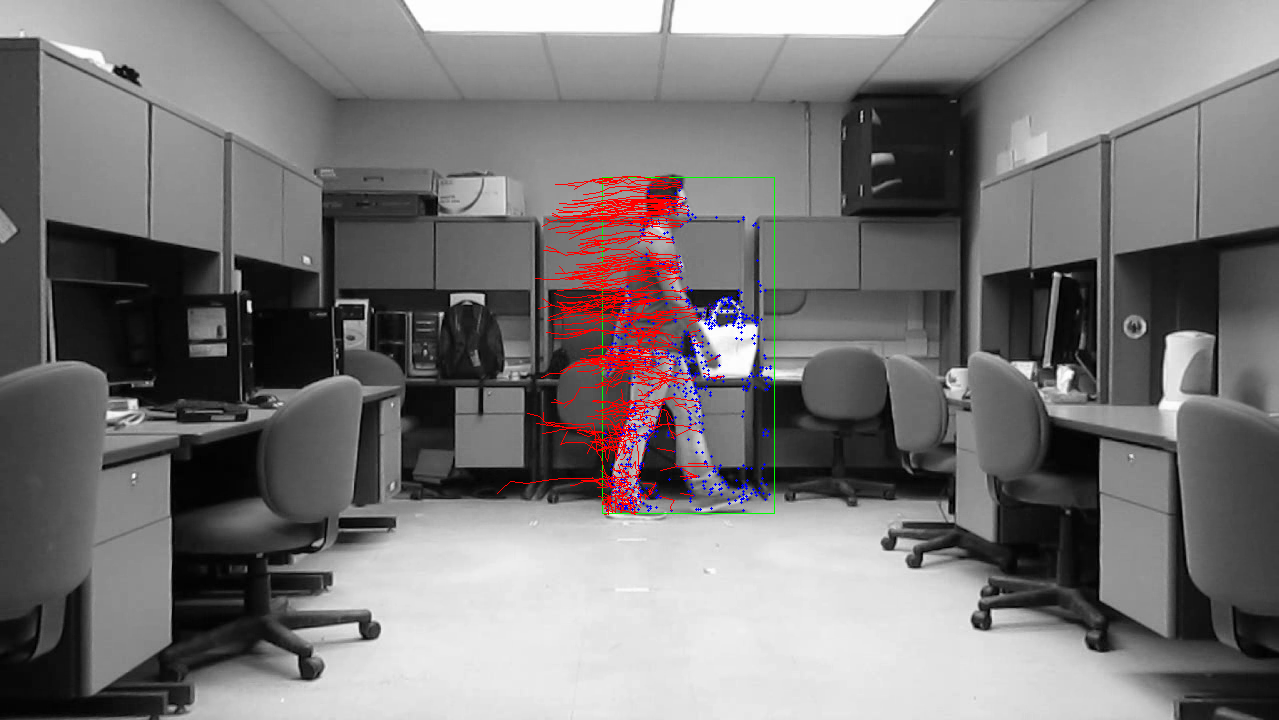}
	\includegraphics[width=0.325\linewidth]{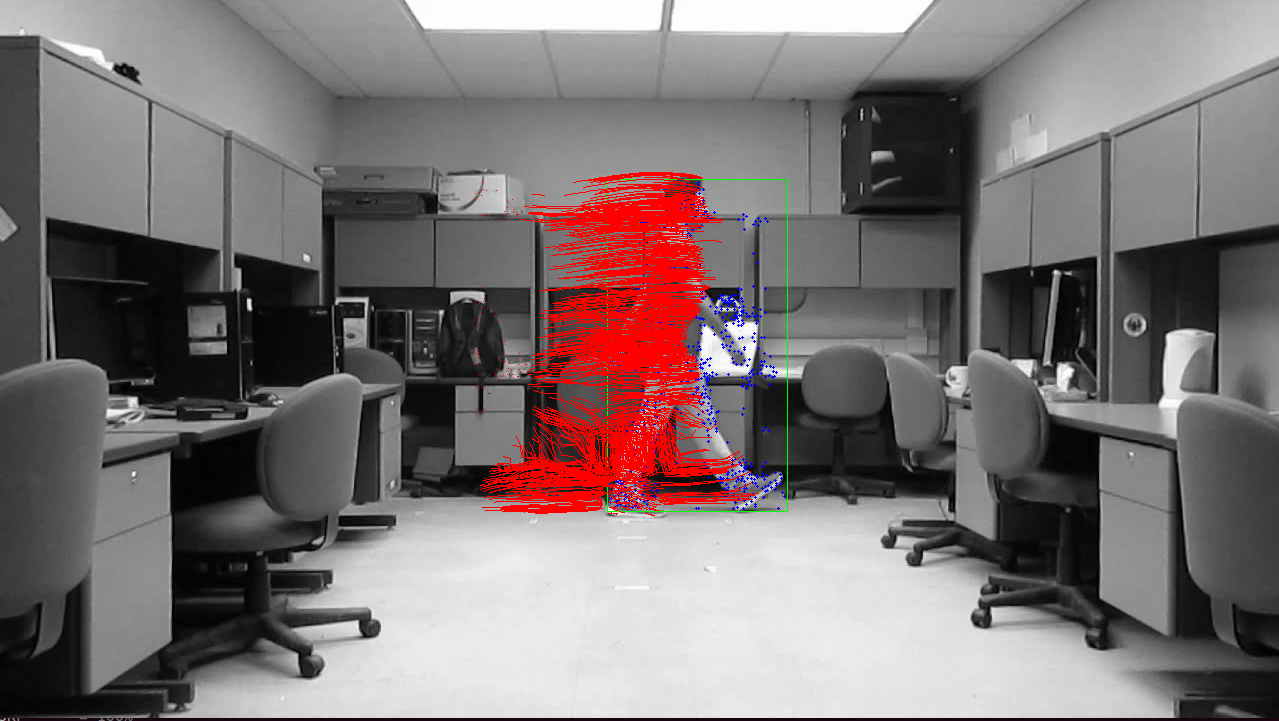}
	\includegraphics[width=0.325\linewidth]{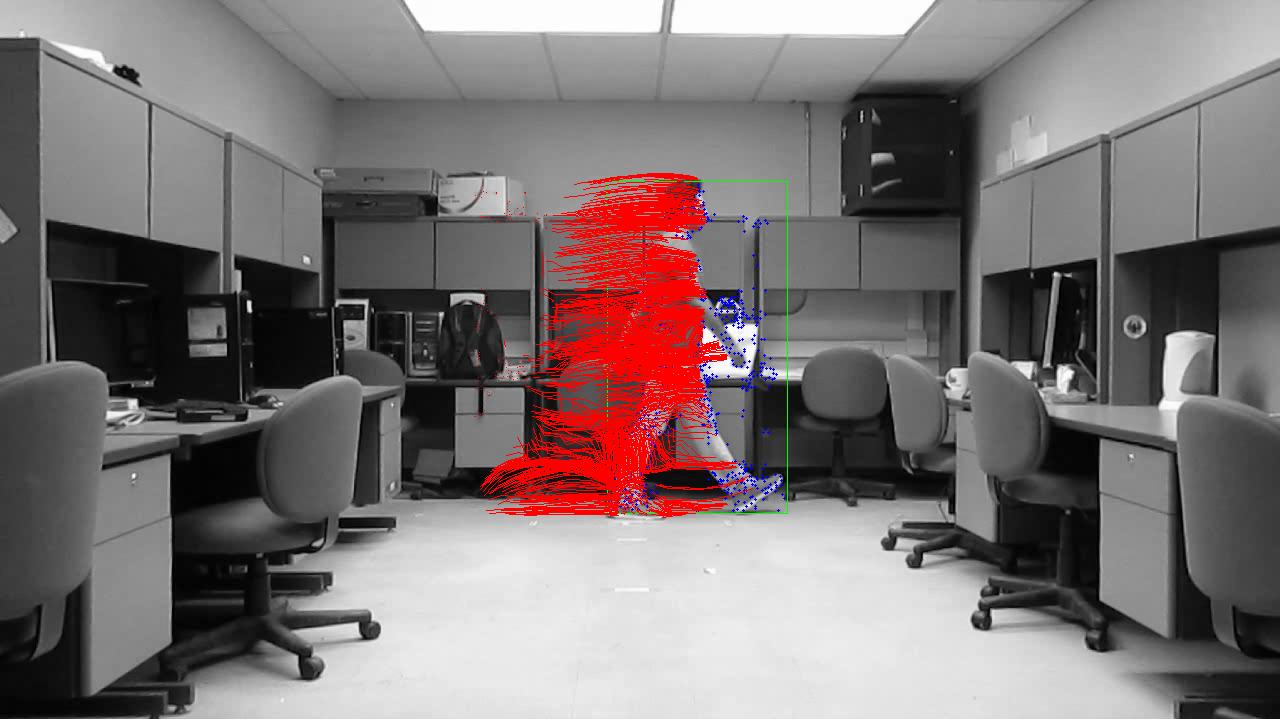}
	\caption{
	2D trajectory extraction algorithms (best seen in color). The 2D trajectories obtained by the three different algorithms: IP (left), LK (middle) and FB (right). The green box is the active area, detected by the preprocessing algorithm.}
	\label{fig::DifferentTrajectoryExtractionAlgorithm}
\end{figure*}

An example of the obtained trajectories is displayed on Figure~\ref{fig::DifferentTrajectoryExtractionAlgorithm} left (best seen in color).

\subsubsection{Lucas-Kanade Feature Point Tracking}
\label{sec::LucasKanadeFeaturePointTracking}
We refer to this method as \textit{``Lucas-Kanade Trajectory''} (LK), as it is based on Lucas-Kanade optical flow algorithm~\cite{bouguet2001pyramidal}. First, the feature points of each frame of the video are extracted. Then, Lucas-Kanade optical flow algorithm is used to find the location of each of these feature points in the next frames, hence creating the trajectories. 

An example of the obtained LK trajectories are displayed on Figure~\ref{fig::DifferentTrajectoryExtractionAlgorithm} (middle).

%

\subsubsection{Farnback Feature Point Tracking}
\label{sec::FarnbackFeaturePointTracking}
This algorithm is similar to the Farnback optical flow algorithm~\cite{farneback2003two}, we refer to as \textit{``Farnback Trajectory''} (FB). Farnback optical flow algorithm is newer than Lucas-Kanade algorithm and has shown a better performance~\cite{wang2011action,farneback2003two}. Farnback optical flow algorithm is also able to provide the dense optical flow field, while Lucas-Kanade optical flow was designed to track sparse feature points. This gives FB an advantage, especially when the selected points are not good feature points. 
However, our tests revealed that FB yields competitive results to LK, and both performed much better than IP.

First, the Farnback optical flow field is calculated before the feature points of each frame are extracted. Starting from the first frame, the location of each point in the next frame is predicted, using the pre-calculated optical flow field. These points are then connected as a trajectory.

The interest points of each frame are extracted using OpenCV implementation of FAST algorithm~\cite{rosten2006machine}. The dense optical flow field is obtained using Farnback motion estimation method~\cite{farneback2003two}. The trajectories created by this algorithm are shown on Figure~\ref{fig::DifferentTrajectoryExtractionAlgorithm} (right). 


\subsection{Disparity-Augmented Trajectory}
\label{sec::3dTrajectoryExtraction}
After the extraction of the 2D trajectories from the left and right videos, matching these trajectories is achieved based on local descriptors (section~\ref{sec::findingMatchingTrajectories}). The matched trajectories are then mapped to their corresponding rectified planes (section~\ref{sec::estimatingTheFundamentalMatrix}). Finally, their disparity is fused with the 2D spatial information (section~\ref{sec::DisparityAugmented}).


\subsubsection{Finding Matching Trajectories}
\label{sec::findingMatchingTrajectories}
Each trajectory starting point, in the left and right videos, are encoded with a SIFT descriptor and the best match of this descriptor is found by using the method in~\cite{lowe2004distinctive}. Starting from the first frame of the video, for each descriptor in the left frame, its best match is found in the right frame. To make the matching robust, we repeat the process between the right to the left frames and, only keep the reciprocal matches.

\subsubsection{Video Rectification}
\label{sec::estimatingTheFundamentalMatrix}
The rectification is the process of mapping an image to a plane, where the y disparity becomes zero and only the x disparity remains. 
If $p$ and $p'$ represent two matching points, between the left and right images, the fundamental matrix $F$ is the matrix that satisfies:

\begin{equation}
\label{equ::fundamentalMatrix}
	pFp' = 0
\end{equation}

The eight-point algorithm is used to estimate $F$~\cite{hartley1997defense} and the FAST algorithm is used to find the feature points. The same algorithm, as explained in Section~\ref{sec::findingMatchingTrajectories}, is used to match the feature points between the left and right video frames. 

\begin{figure*}[!htbp]
\centering

\begin{tikzpicture}
\begin{axis}[
    width = 0.5\linewidth,
    xlabel={Sample stereo trajectories},
    legend style={at={(0cm,2cm)},anchor=north west},
    ymajorgrids=true,
    grid style=dashed,
]

\addplot[color=red, mark=*] 
	coordinates {(0.529282, 0.302554) (0.536277, 0.30275) (0.543578, 0.302267) (0.550729, 0.301873) (0.557384, 0.300724) (0.564094, 0.299299) (0.570512, 0.297437) (0.576743, 0.295539) (0.582921, 0.294298) (0.589273, 0.294611) (0.596294, 0.296409) (0.603279, 0.298644) (0.610352, 0.300977) (0.617153, 0.301959) (0.624323, 0.303037) (0.630893, 0.30179) (0.637772, 0.301413) (0.644227, 0.300913) (0.650214, 0.299864)};

\addplot[color=blue, mark=triangle] 
	coordinates {((0.497238, 0.35167) (0.50393, 0.351721) (0.511314, 0.351311) (0.518454, 0.350931) (0.525235, 0.349637) (0.531665,0.347923) (0.53786, 0.345613) (0.543955, 0.34366) (0.550198, 0.342668) (0.55675, 0.343134) (0.563674, 0.34486) (0.5706, 0.347153) (0.577617, 0.349838) (0.584726, 0.351184) (0.592069, 0.352816) (0.598828, 0.352313) (0.605394, 0.35226) (0.611757, 0.351388) (0.617767, 0.350127)
};



\legend{Left,Right}
\end{axis}
\end{tikzpicture}
\begin{tikzpicture}
\begin{axis}[
    width = 0.5\linewidth,
    xlabel={Sample rectified stereo trajectories},
    legend pos=south east,
    ymajorgrids=true,
    grid style=dashed,
]

\addplot[color=red, mark=*] 
	coordinates {(4.083905696505915062e-01, 4.338906762270391959e-01) (4.128460923664530324e-01, 4.303566863055560687e-01) (4.175455350808225385e-01, 4.278649741712325372e-01) (4.228130121295593513e-01, 4.252452342611746650e-01) (4.281978586783060092e-01, 4.231155667713080160e-01) (4.338024100564202845e-01, 4.220212101234350754e-01) (4.396758753686350607e-01, 4.223456097583975222e-01) (4.460441637970914419e-01, 4.241732163734988625e-01) (4.526213828738526179e-01, 4.262759723642053356e-01) (4.592340856599764765e-01, 4.291889710551127779e-01) (4.661516333075107488e-01, 4.323922498836437756e-01) (4.739601546342224192e-01, 4.351021953978409518e-01) (4.819474846900381193e-01, 4.361141321906819024e-01) (4.894330312493821999e-01, 4.361541489028571972e-01) (4.971266800482095904e-01, 4.360592967061021619e-01) (5.040249398043920026e-01, 4.338208203460862666e-01) (5.108791425560605948e-01, 4.310740016461834800e-01) (5.172717852087800328e-01, 4.285037086127621575e-01) (5.238387279061120783e-01, 4.269142257463990897e-01)};

\addplot[color=blue, mark=triangle] 
	coordinates {(4.126085621647145318e-01, 4.383011120326761323e-01) (4.170200878163817793e-01, 4.349161560264208637e-01) (4.215291028610345658e-01, 4.323320438445366620e-01) (4.268009257073381302e-01, 4.297468619085484387e-01) (4.320070069640152832e-01, 4.270153201053751846e-01) (4.374476680780373661e-01, 4.256893622602431759e-01) (4.431303032820738119e-01, 4.262406073689005703e-01) (4.493344927906778308e-01, 4.281324064677089480e-01) (4.558125279932305207e-01, 4.306005470739864283e-01) (4.623065813851148786e-01, 4.334953806326550252e-01) (4.695988180261841949e-01, 4.372111220243641294e-01) (4.771945412856147817e-01, 4.399158889782136539e-01) (4.848215927314672991e-01, 4.403503725925374535e-01) (4.929357581106532105e-01, 4.410013659359509064e-01) (5.006954567535094602e-01, 4.403354539590202399e-01) (5.081466119099244949e-01, 4.389694762542205586e-01) (5.149207272212690256e-01, 4.353305780036445527e-01) (5.217271873509317093e-01, 4.332055187301791510e-01) (5.283822536783573387e-01, 4.317608699122651439e-01)};



\legend{Left,Right}
\end{axis}
\end{tikzpicture}

\caption{
Sample stereo trajectories, unrectified (left) and rectified (right). (best seen in color)
 }	
\label{fig::trajectorySamples}
\end{figure*}
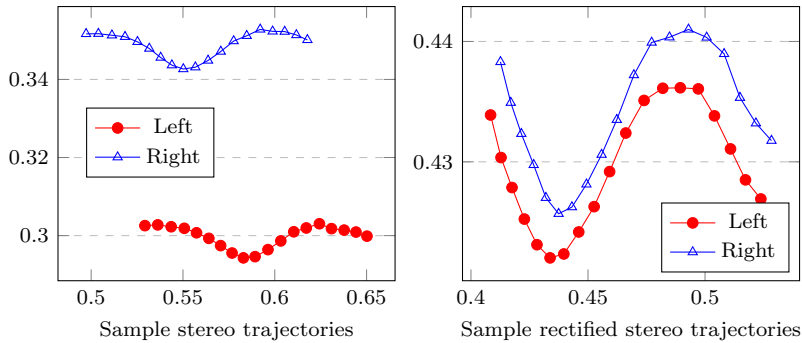

The calculations of $F$ and rectification matrices $H_l$ and $H_r$~\cite{hartley1999theory} depend on the quality of matched points. To address these issues, we propose the following technique to find the best estimation of $F$, $H_l$ and $H_r$. First, $m$ random frames of the stereo video are selected. For each pair $i$ of these stereo frames, $F^i$, $H_l^i$ and $H_r^i$ are calculated. If $p=(x,y,1)^T$ and $p'=(x',y',1)^T$ represent a matching point, then $q=H_l^ip=(u,v,1)^T$ and $q'=H_r^ip'=(u',v',1)^T$ represents the mapping of these corresponding points on the rectified plane, where ideally $v - v' = 0$. Considering the matched trajectories from section~\ref{sec::findingMatchingTrajectories}, the best estimate of $F$ is the one that maximizes the number of trajectories that will be rectified with an acceptable y disparity. Figure~\ref{fig::trajectorySamples} shows samples of two matching trajectories, one before rectification and the other one after rectification.
In addition, because the calculation of $F$ is susceptible to outliers, we 
have also used the random sample consensus (RANSAC) method to make its 
calculation robust.


\subsubsection{Calculating Disparity-Augmented Trajectories}
\label{sec::DisparityAugmented}
Having the rectification matrices $H_l$ and $H_r$, it is now easier to calculate the rectified left and right trajectories, $t_l$ and $t_r$, as

\begin{align}
	t_l &= H_l T_l \\
	t_r &= H_r T_r
\end{align}
where $T_l$ and $T_r$ are the homogeneous representation of the left and right trajectories, respectively. 

Each column of $t_l$ or $t_r$ represents a rectified trajectory image plane point. Consider two corresponding points $q=(u,v,1)^T$ and $q'=(u',v',1)^T$ on left and right trajectories, respectively. The corresponding disparity 
augmented point $(x_m,y_m,d)$ will be given by:

\begin{align}
	x_m &= u   \\
	y_m &= v	 \\
	d &= u - u' 
\end{align}

\section{Trajectory Shape Descriptor}
\label{sec::trajectoryShapeEncoding}
The method we used to create a descriptor for trajectories is an extension of~\cite{habashi2017better}, where the locations of interest points in 2D or 3D (2D plus disparity) spaces are encoded into trajectories. The latter are considered as discrete functions, that map time values to coordinates. The first and second derivatives of such mapping function, with respect to time, represent respectively the velocity and the acceleration of the interest point. Higher order derivatives encode higher order motion information. The final descriptor is obtained by concatenating these derivatives. In our experiments, we have used derivatives up to the 7th order, for single views and up to 5th order for multiple views.

Formally, each trajectory, defined in Equation~\ref{equ::Trajectory}, can be interpreted as a function of time that map time values to locations in $\mathbb R^{n}$ space.

\begin{align}
	\label{eq::Trajectory}
		&T : time \to \mathbb R^{n} \\ 
		&T(t) = P(\frac{t - t_0} {\Delta t}) \\
		&P(i) = p_i \in \mathbb R^{n}
\end{align}
where $\Delta t$ is the time between two consecutive frames and $t_0$ is the starting time of trajectory $T_k$. 

The first derivative of this function is given by:

\begin{equation}
      \label{eq::Differentiation}
      V(t) = \frac{dT}{dt} = \lim_{t \to 0} \frac{T(t+dt) - T(t)}{dt}
\end{equation}

Because a video typically has a fixed frame rate, the smallest value of $dt$ is the time distance between two consecutive frames ($\Delta t$). Changing the unit of measurement from seconds to frames makes $dt = 1$. Hence, the velocity equation~\ref{eq::Differentiation} can be rewritten as: 

\begin{align}
\label{equ::velocityDescriptor}
	V(t) &= T(t+1) - T(t) \\
	V&= (v_0, v_1, ..., v_{l-1}) \\
	v_i&=p_{i+1} - p_{i}, i=0...{(l-2)}
\end{align}

Similarly, the acceleration is given by

\begin{align}
	\label{eq:Acceleration}
	A(t) &= \frac{dV(t)}{dt}= V(t+1) - V(t) \\
	A&= (a_0, a_1, ..., a_{l-1}) , \\
	a_i&=v_{i+1} - v_{i}, i=0...{(l-2)}
\end{align}

Higher order functions can be defined as follows:

\begin{equation}
	\label{eq::HigherOrderDrivitives}
	\nabla^{(n)} = \frac{d^nT(t)}{dt^n} 
\end{equation}

The actual descriptor is created by the concatenation of these derivatives. For example, descriptor $D_1$ is given by the velocity ($\nabla^{(1)}$), descriptor $D_2$ is given by the concatenation of $\nabla^{(1)}$ and $\nabla^{(2)}$ and, descriptor $D_3$ is given by the concatenation of $\nabla^{(1)}$, $\nabla^{(2)}$ and $\nabla^{(3)}$.

\section{Learning phase}
\label{sec::learningPhase}
Our proposed trajectory shape representation, like other sparse representation methods, represents a video by a set of independent features. Formally, a video can be represented by a set of feature descriptors as:
\begin{equation}
  S=\{D_k| D_k \in \mathbb{R}^N\}
\end{equation}

where $N$ is the dimension of the local descriptors. 


Existing machine learning methods in general and SVM in particular, expect data as a vector of predetermined size. As a result, each set of these features should be represented by a vector. Different methods have been proposed in the literature. 
One of the conventional methods to convert sparse sets to a vector is based on the bag of words (BOW)~\cite{pang2002thumbs}\cite{zhang2010understanding}. 
Another favorite technique, known as Fisher Vector Encoding (FVE)~\cite{perronnin2007fisher,perronnin2010improving}, combines the generative and discriminative methods~\cite{varol2015efficient}. Unlike BOW that uses only the first order statistics, FVE uses first and second order statistics for encoding~\cite{wang2013action}. Instead of using K-Means for clustering, Expectation Maximization (EM) is used to cluster data into K Gaussian Mixtures. Then, the created Gaussian mixture model (GMM) is used to estimate the means, variances and prior probabilities of the mixtures.

%
%
%
%
%



\section{Results and discussion}
\label{sec::experiments}
This section first provides details on the used datasets and on our experimental setup.

\subsection{Dataset}
\label{sec::dataset}

\begin{figure*}[!htbp]
	\centering
	\includegraphics[width=0.41\linewidth]{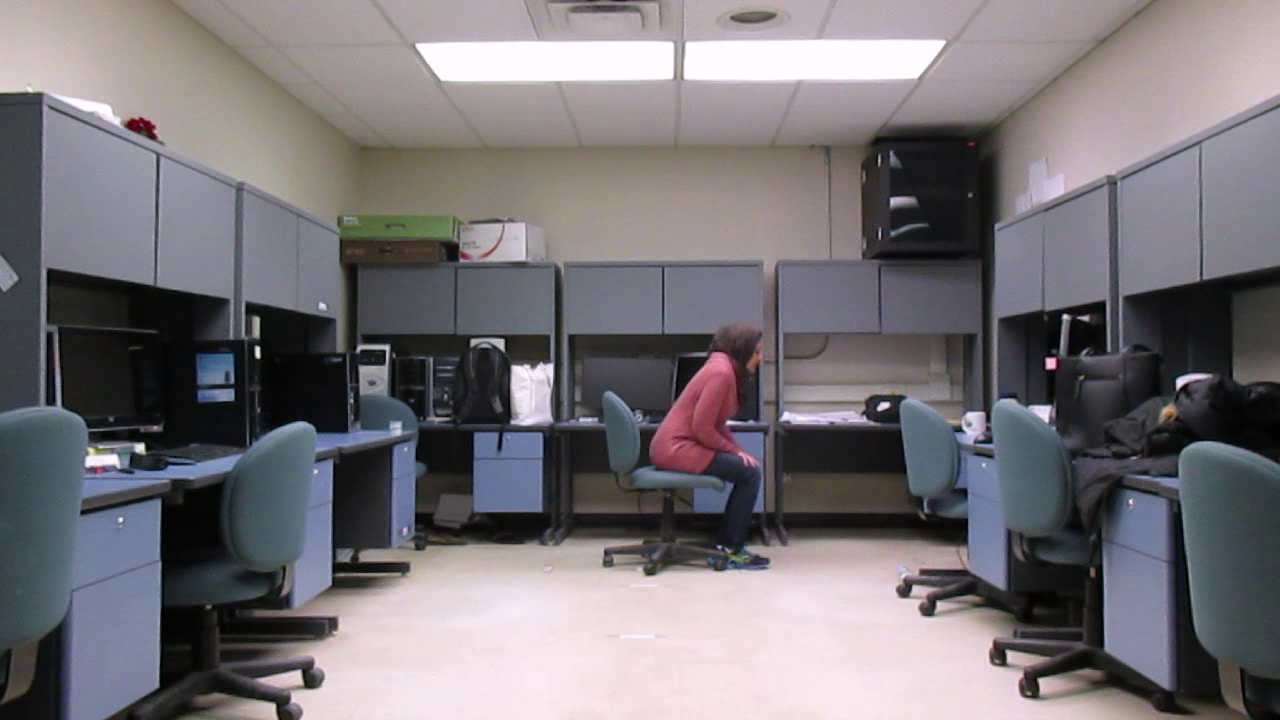}
	\includegraphics[width=0.41\linewidth]{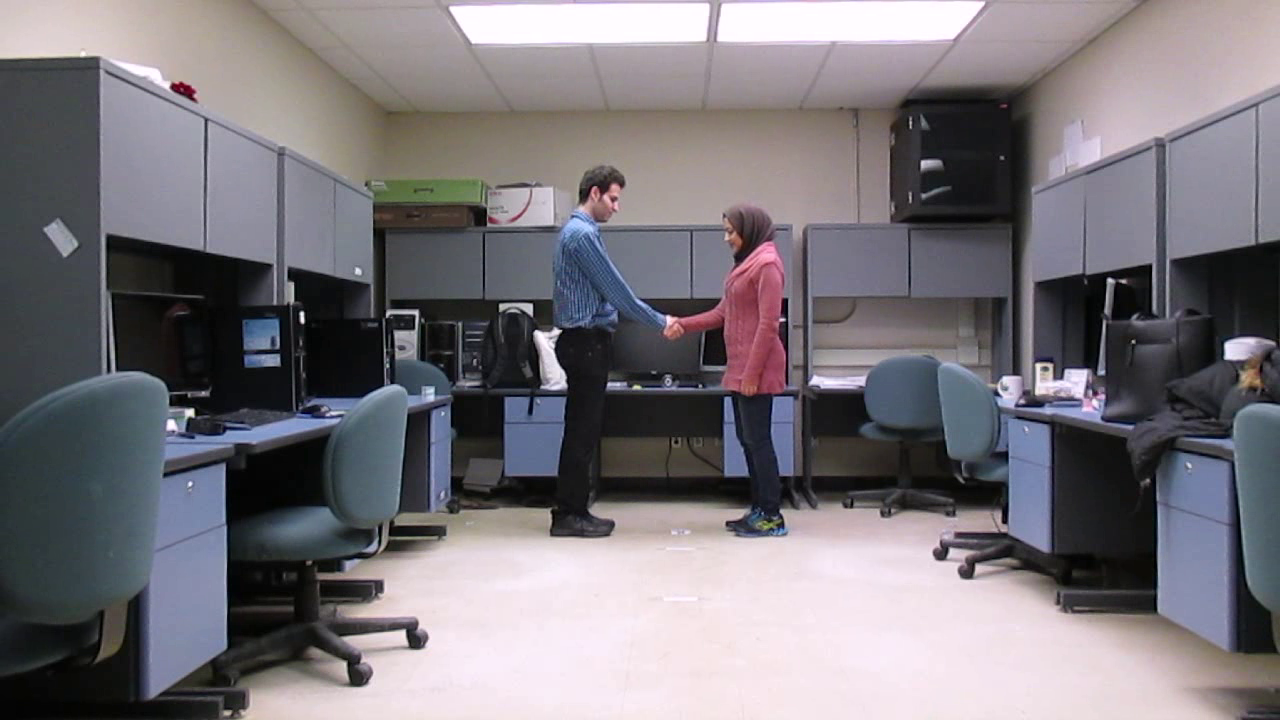}

	\includegraphics[width=0.41\linewidth]{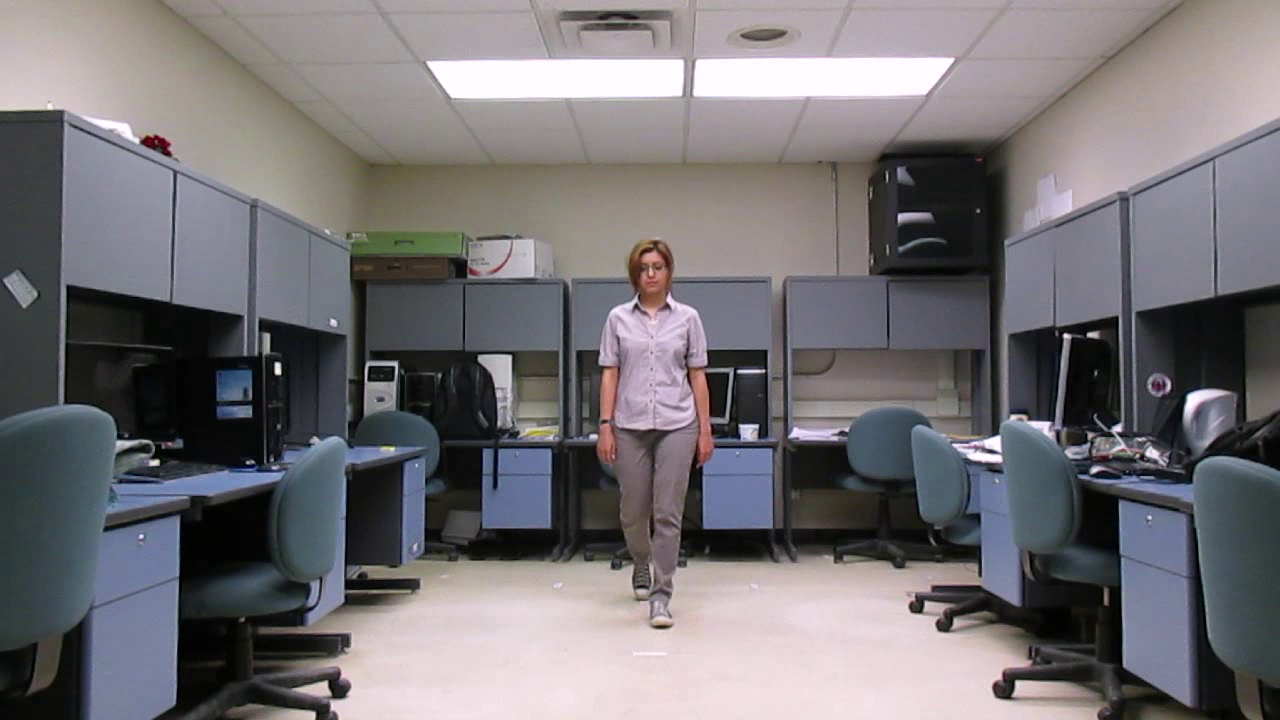}
	\includegraphics[width=0.41\linewidth]{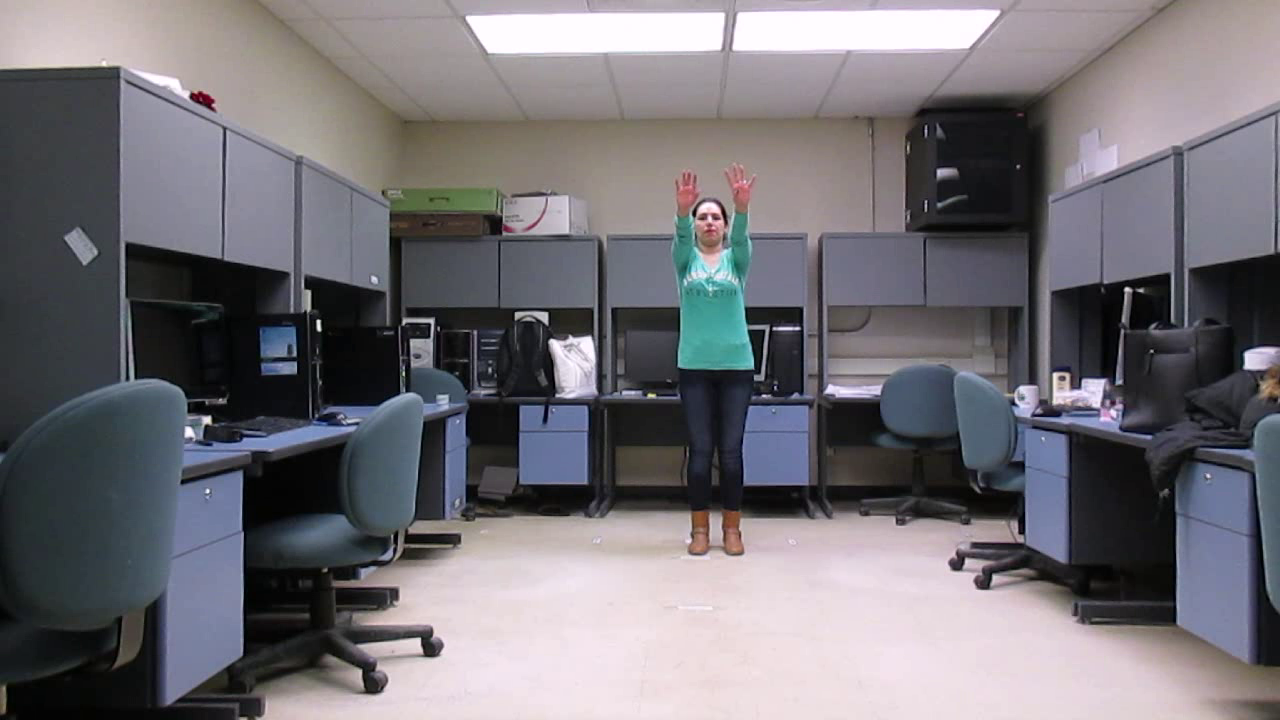}
	
	\includegraphics[width=0.41\linewidth]{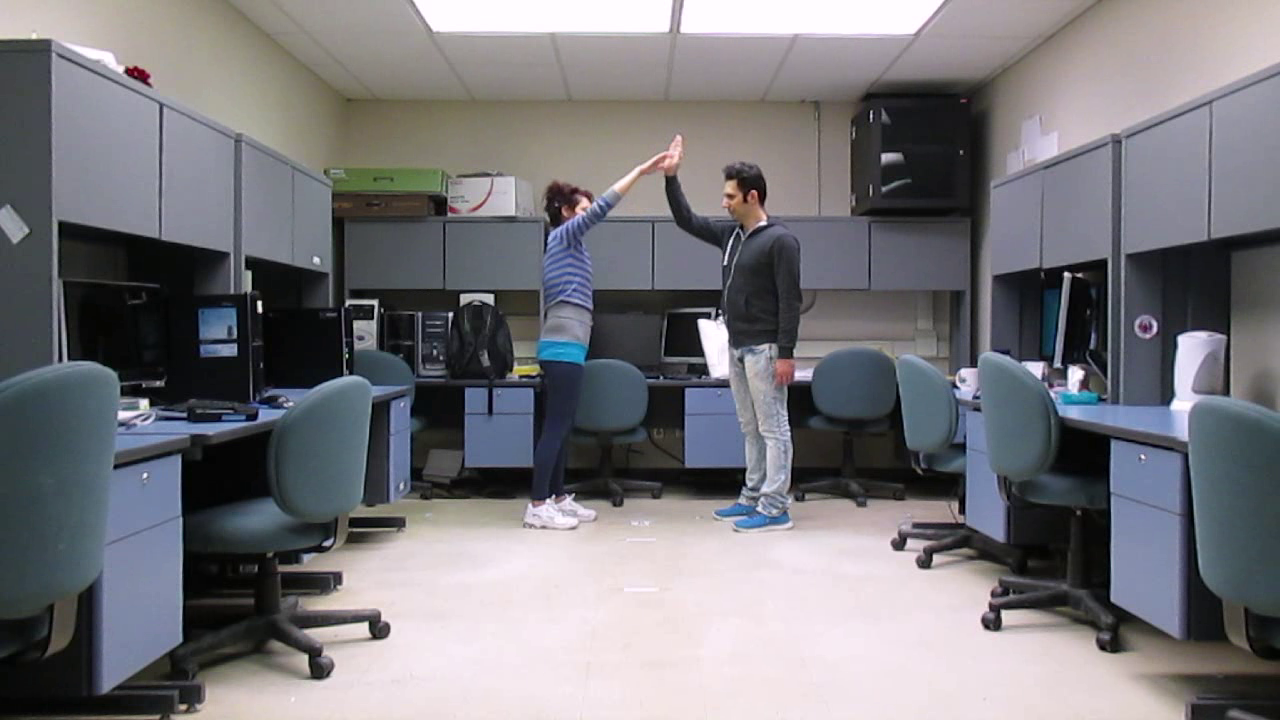}
	\includegraphics[width=0.41\linewidth]{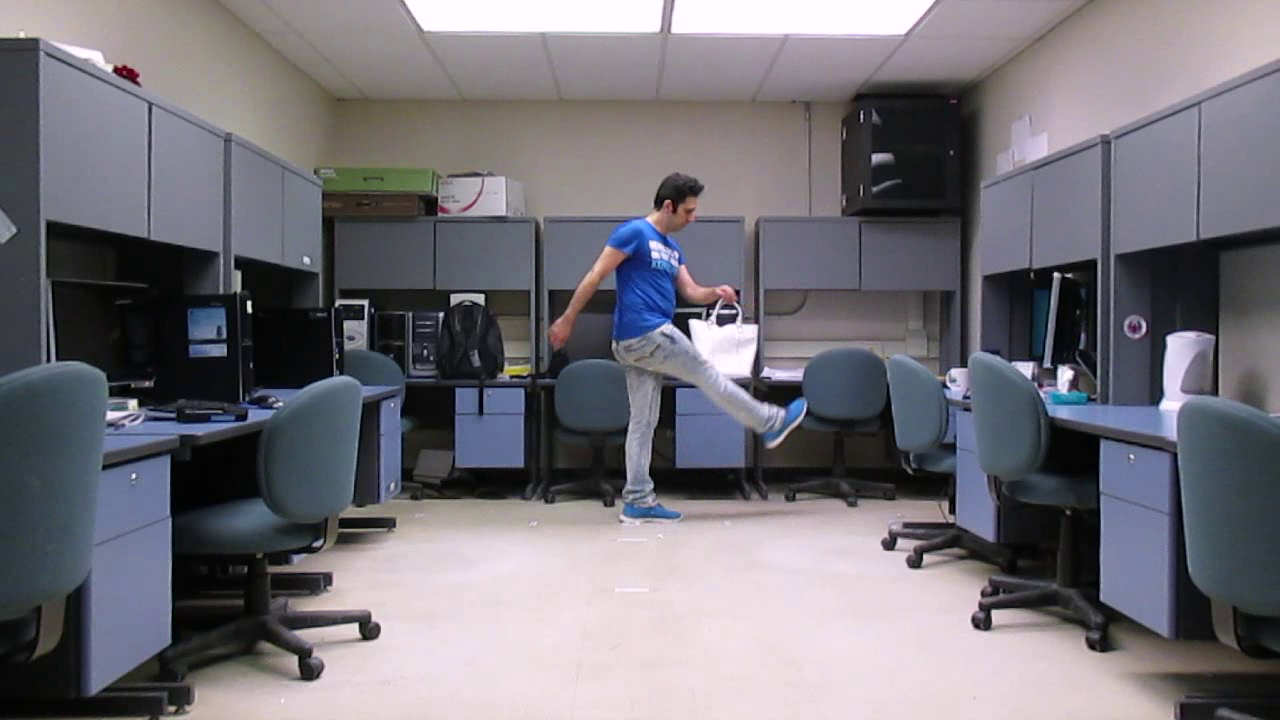}
	
	\includegraphics[width=0.41\linewidth]{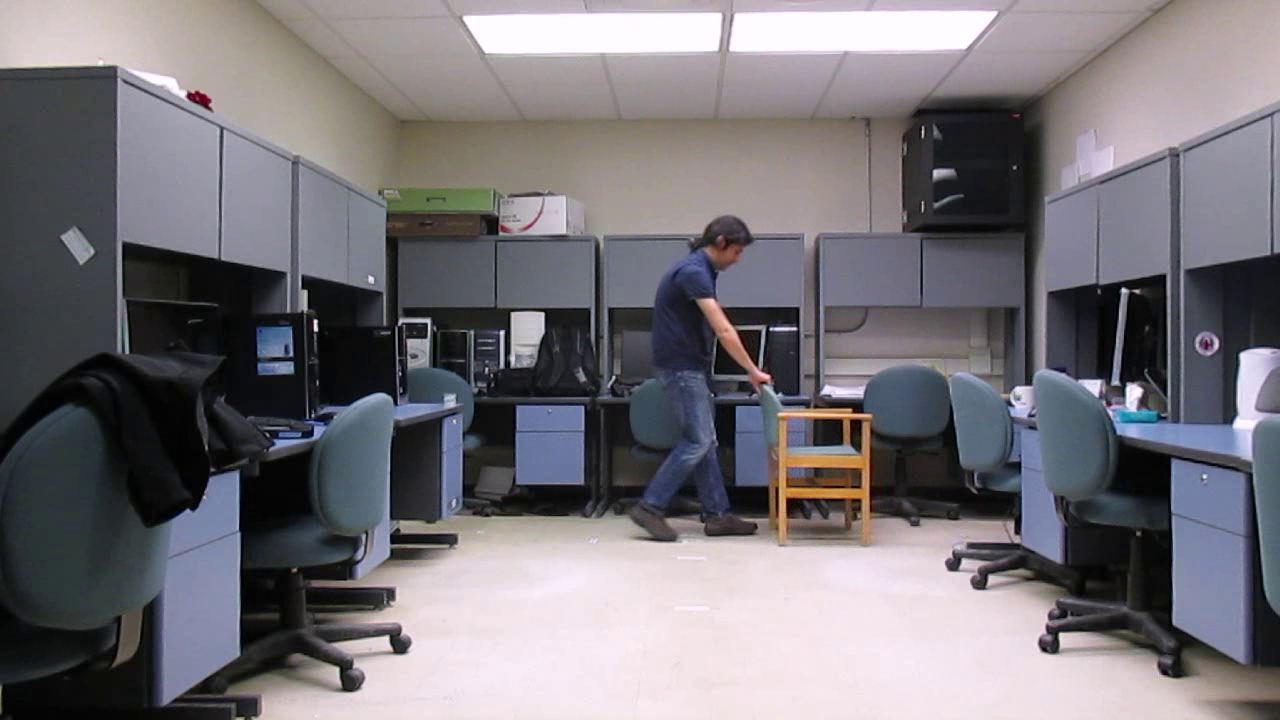}
	\includegraphics[width=0.41\linewidth]{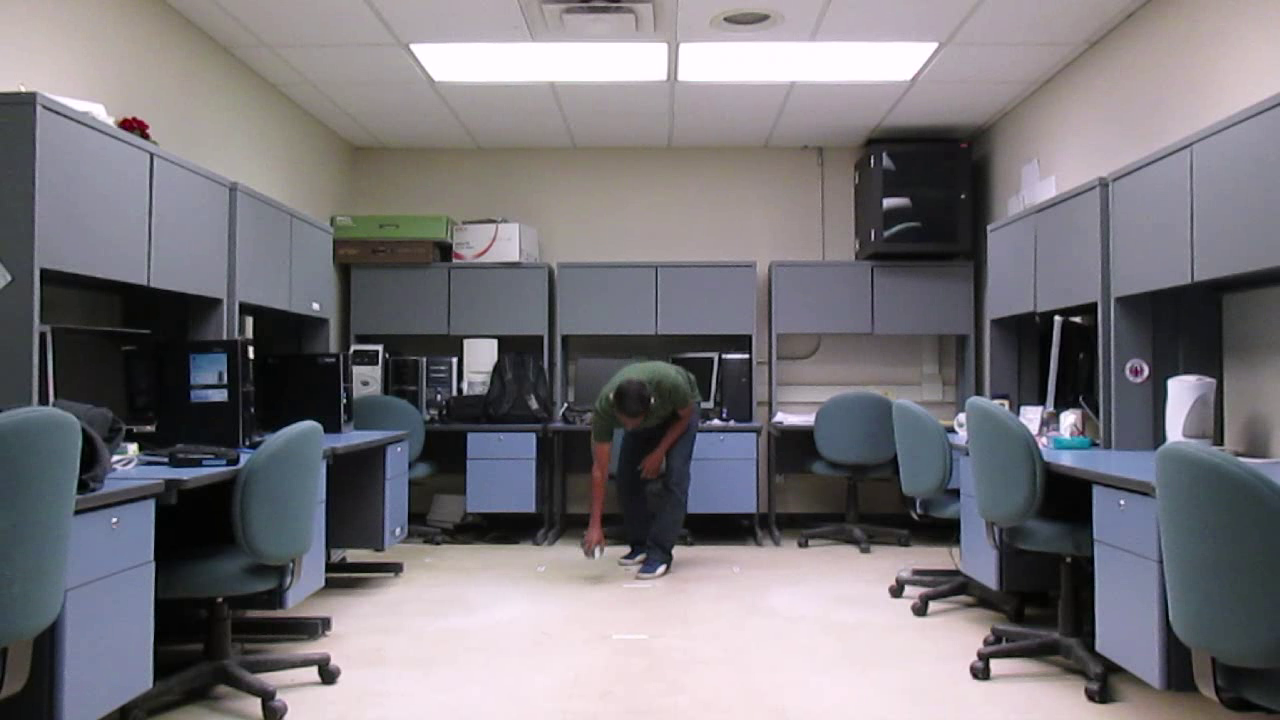}
		
	\includegraphics[width=0.41\linewidth]{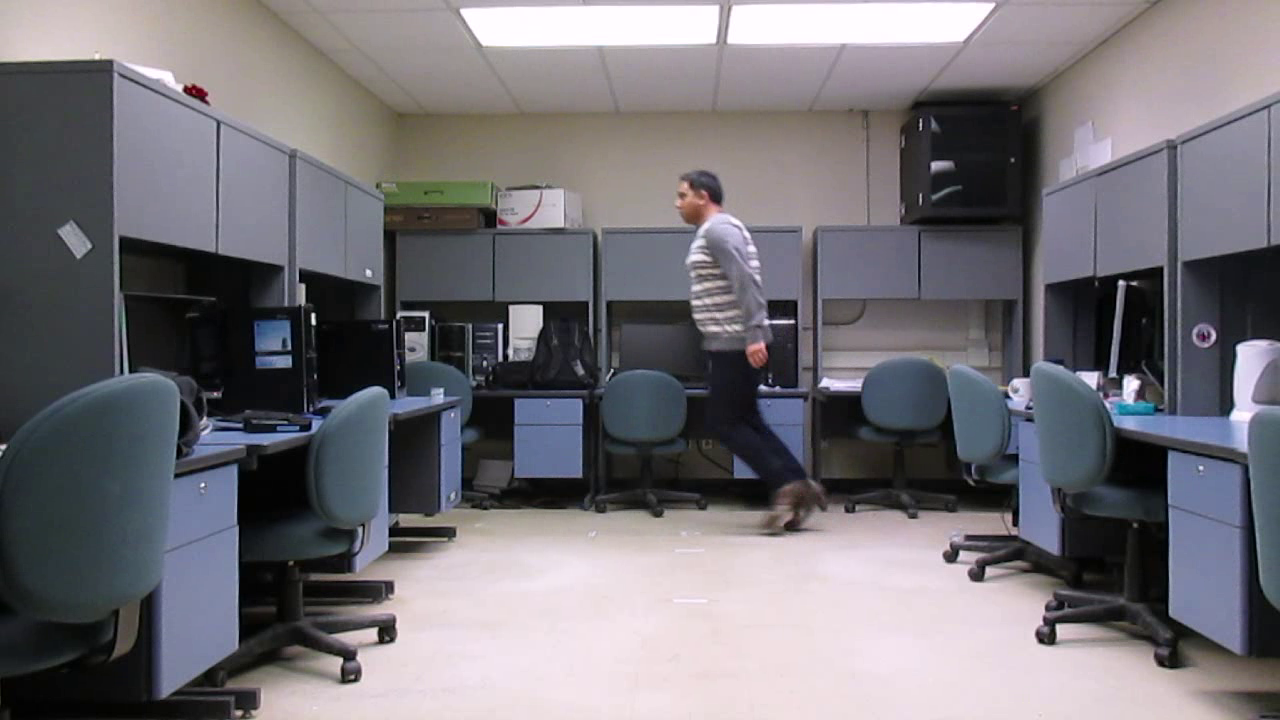}
	\includegraphics[width=0.41\linewidth]{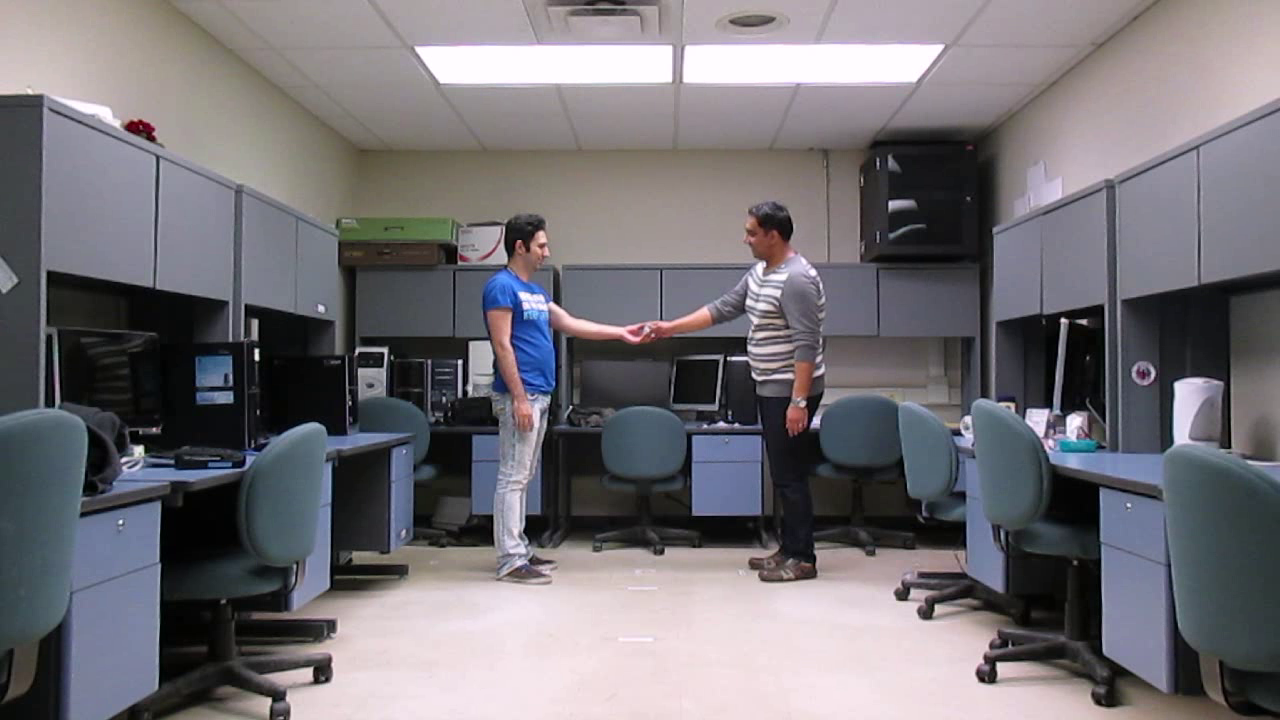}
	
	\includegraphics[width=0.41\linewidth]{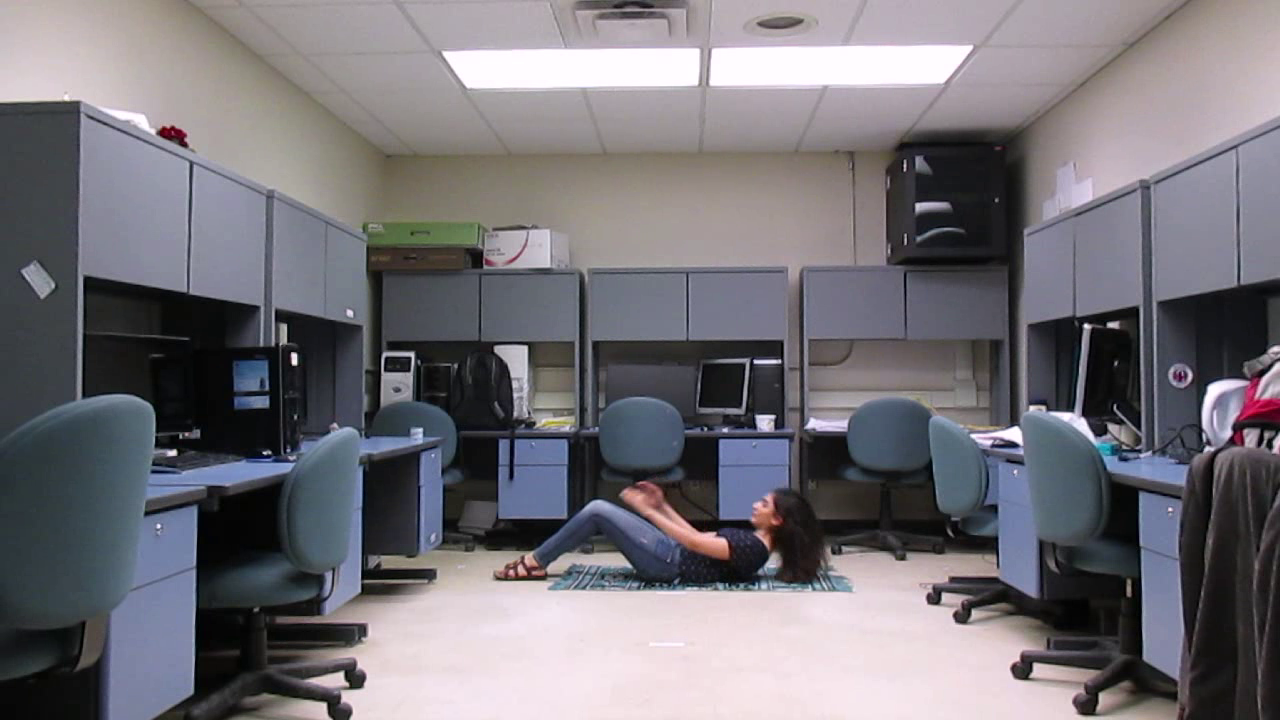}
	\includegraphics[width=0.41\linewidth]{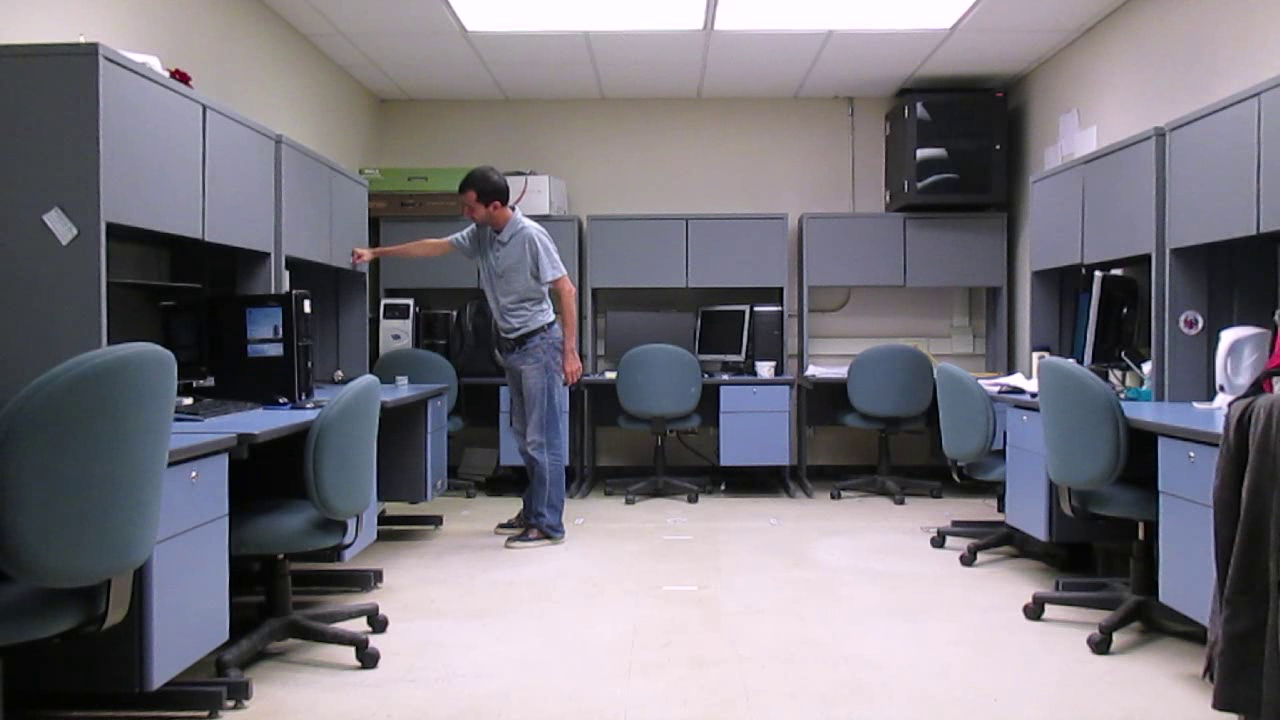}
	
	\caption{
	Sample images from our dataset (best seen in color) First row: standing up, hand shaking, walking toward camera and two-hand waving. Second row: high-five, shooting the ball, pulling heavy object and picking up something from floor. Third row: jumping over gap, exchanging object, sitting up and punching 
}
\label{fig::sampleImagesFromDataset}
\end{figure*}

To the best of our knowledge, the only existing stereo dataset is the Hollywood 3D dataset, which has 13 different activity classes and a special "No Action" class. 
This dataset is a very challenging one for trajectory-based methods, as the lengths of some videos are too short, making it impossible to create meaningful trajectories. In addition, sporadic camera motions create unwanted trajectories, which degrades the accuracy. There are also several videos that are labeled as single activities, while other activities are going on in the background. This usually confuses methods that assume a single activity in the field of view.

We have also created our own stereo dataset with 27 different activity classes, using static cameras. 


Note that despite the amount of work that has been done on human action recognition, there is no universally accepted definition of a human action. 
This is especially visible in the different datasets that have been created so far. 
Some actions, like walking, running and jumping, are widely accepted~\cite{schuldt2004recognizing,zelnik2001event}. A single person usually performs these activities, with some of them containing human and objects/environment interactions, like riding a bike or shooting a basketball~\cite{liu2009recognizing}, playing cello and mopping the floor~\cite{soomro2012ucf101}. Some researchers went beyond this and created datasets for cooking different recipes~\cite{rohrbach2012database}.



In our own dataset, we considered actions to be sole movements of the human body, regardless of the background, environment or tools they might be using. Besides, we studied activities that contain the whole human body movement as actions. We considered 27 different actions for this dataset. The activities were selected based on the frequency of their appearance in other human datasets. Besides, we added activities that can be performed in an office setting. We did not use  tools a lot. For example, for recording throwing, the actor does not throw anything, he/she simply acts like throwing. The only exception was for pushing/pulling objects, where a chair have been used. Our actions were performed by eleven different volunteers in an everyday office setting. Some activities were performed in different scenarios. For example, walking was performed four times by each actor: walking from left to right, walking from right to left, walking toward the camera and walking away from the camera. We used off-the-shelf cameras to record these activities. Our stereo system consisted of two cameras that were roughly at 30cm from each other.
Five hours of activities were recorded by each camera, allowing us to obtain 4076 stereo video clips, from which 1188 were selected to represents 27 different activities. Each activity was performed four times by each of the eleven actors. Table~\ref{tab::activityNames} presents information about the recorded activities and some samples of these activities are shown on Figure~\ref{fig::sampleImagesFromDataset}.

\begin{table}
\caption{The list of activity classes in our own dataset}
\label{tab::activityNames}
\centering
\begin{tabular}{ccc}
\toprule
Crossing Arms       & Exchange Object      & Hand Clapping \\
Hand Shaking        & Hand Waving          & High Five     \\
Hitting             & Jumping Over Gap     & Jumping Jack  \\
Kick the Ball       & Kicking              & Lay Down      \\
Pickup(Floor)       & Pull                 & Pointing      \\
Pickup(Table)       & Push                 & RaiseHand     \\
Running             & Scratch Head         & Sit down       \\
Sit-up               & Skipping             & Standup       \\
Throwing            & Turning              & Walking       \\
\bottomrule
\end{tabular}

\end{table} 


\subsection{Experimental Setup} 
Our tests were carried out on a Ubuntu machine, with eight 3.8 GHz cores and 8Gb of RAM. The video processing part, including trajectory extraction, was implemented in C++, using OpenCV library. The trajectory aligning algorithm was implemented in Python.

After obtaining a set of trajectory descriptors for each video, and since \cite{wang2013action} has shown the `effectiveness' of \emph{Fisher Vectors} over other methods, we have used  \emph{Fisher Vectors} to prepare data before passing it to a standard support vector machine (libSVM~\cite{CC01a}).

The data used for training and testing was split as follows. For each action, all videos of one actor are used for testing, while the remaining videos from other actors are used for training. A confusion matrix is calculated for each 
action. The blending of these matrices represents the overall confusion matrix. The accuracy, reported in the tables, is the ratio of correctly classified instances to the total number of samples, directly calculated from the overall confusion matrix.

\subsection{2D Trajectories}
\label{sec::results}

\begin{table*}

	\centering
\caption{The accuracy(\%) measured for left, right and both cameras. Both-camera is the combination of features from left and right frames}
	\label{tab:IP_LK_FB_TrajectoryLength}
\begin{tabular}{cccccccccc}
\toprule	
		& \multicolumn{3}{c}{Left Camera} &   \multicolumn{3}{c}{Right Camera}    &   \multicolumn{3}{c}{Both Cameras}      \\
	\cmidrule(lr){2-4} \cmidrule(lr){5-7} \cmidrule(lr){8-10}
	Length  &   IP    &   LK   &   FB   &   IP    &   LK    &   FB    &   IP    &   LK    &    FB     \\
	\midrule                                                                                            
	9 	& 77.73 & 83.87 & 82.81 & 74.75 & 81.09 & 82.92 &80.20 & 82.89 &82.97 \\
	11 	& 76.55 & 84.29 & 84.50 & 74.45 & 84.62 & 83.38 &\textbf{80.81}  &82.49 & 85.98 \\
	13	& \textbf{79.12}  & 84.71 & 85.72 & 73.54 & 82.44 &85.80 & 79.54 &83.96 & 85.04 \\
	15	& 77.68 & 86.06 & 84.37 & 72.72 & 85.04 & 84.58 &79.80 & 84.97 &86.78 \\
	17	& 78.93 & 84.54 & 84.54 & 73.54 & 86.54 & 85.31 &78.09 & 86.07 &86.08 \\
	19	& 77.33 & 85.13 & 85.46 & 73.77 & 87.12 & 86.25 &79.78 & 85.90 &87.05 \\
	21	& 77.22 & 86.74 & 85.88& \textbf{75.67} & 85.66 & \textbf{87.90} &79.45 & \textbf{86.82} & \textbf{88.19} \\
	23	& 77.42 &  \textbf{86.81} &  \textbf{87.90} &  73.34  & \textbf{87.29}  &  87.64  &  78.70  &  85.52  &   87.43   \\
	25	&  78.94  &  86.22 &  87.65 &  71.33 &  84.94  &  85.37  &  79.74 &  85.64  &   87.04  \\
	27	&  75.32 &  86.69 &  85.84 &  70.37 &  86.52 &  87.70 & 77.85  &  85.18  &   87.35   \\
	\bottomrule
\end{tabular}
\end{table*}

\begin{figure*}[!htbp]
	\centering
\begin{tikzpicture}
\begin{axis}[
    width = 0.33\linewidth,
    title={Left Camera},
	xlabel={Trajectory length [frames]},
    ylabel={Accuracy [\%]},
    xmin=11, xmax=27,
    ymin=70, ymax=95,
    xtick={5,10,15,20,25},
    legend style={at={(-0.5cm,4cm)},anchor=north west},
    ymajorgrids=true,
    grid style=dashed,
] 
\addplot[
    color=blue,
    mark=square*,
    ]
    coordinates { (5,  75.79)(7,  75.03) (9,  77.73) (11, 76.55) (13, 79.12) (15, 77.68) (17, 78.93) (19, 77.33) (21, 77.22) (23, 77.42) (25, 78.94) (27, 75.32)};
\addplot[
    color=red,
    mark=*,
    ]
    coordinates { (5,  77.73) (7,  81.68) (9,  83.87) (11, 84.29) (13, 84.71) (15, 86.06) (17, 84.54) (19, 85.13) (21, 86.74) (23, 86.81) (25, 86.22) (27, 86.69) 
    };                                                                
\addplot[
    color=green,
    mark=triangle*,
    ]
    coordinates { (5,  71.25) (7,  78.90) (9,  82.81) (11, 84.50) (13, 85.72) (15, 84.37) (17, 84.54) (19, 85.46) (21, 85.88) (23, 87.90) (25, 87.65) (27, 85.84) };                                                                
    \legend{IP, LK, FB}                                                       

\end{axis}                                                            
\end{tikzpicture}
\begin{tikzpicture}
\begin{axis}[
    width = 0.33\linewidth,
    title={Right Camera},
	xlabel={Trajectory length [frames]},
    xmin=11, xmax=27,
    ymin=70, ymax=95,
    xtick={5,10,15,20,25},
    legend pos=north west,
    ymajorgrids=true,
    grid style=dashed,
]
 
\addplot[
    color=blue,
    mark=square*,
    ]
    coordinates { (5,  73.69) (7,  71.62) (9,  74.75) (11, 74.45) (13, 73.54) (15, 72.72) (17, 73.54) (19, 73.77) (21, 75.67) (23, 73.34) (25, 71.33) (27, 70.37) };
\addplot[
    color=red,
    mark=*,
    ]
    coordinates { (5,  73.77) (7,  77.47) (9,  81.09) (11, 84.62) (13, 82.44) (15, 85.04) (17, 86.54) (19, 87.12) (21, 85.66) (23, 87.29) (25, 84.94) (27, 86.52) };                                                                
\addplot[
    color=green,
    mark=triangle*,
    ]
    coordinates { (5,  72.55) (7,  80.92) (9,  82.92) (11, 83.38) (13, 85.80) (15, 84.58) (17, 85.31) (19, 86.25) (21, 87.90) (23, 87.64) (25, 85.37) (27, 87.70) };                                                                

\end{axis}                                                            
\end{tikzpicture}
\begin{tikzpicture}
\begin{axis}[
    width = 0.33\linewidth,
    title={Both Cameras},
	xlabel={Trajectory length [frames]},
    xmin=11, xmax=27,
    ymin=70, ymax=95,
    xtick={5,10,15,20,25},
    legend pos=north west,
    ymajorgrids=true,
    grid style=dashed,
]
 
\addplot[
    color=blue,
    mark=square*,
    ]
    coordinates {(5,  72.62) (7,  78.25) (9,  80.20) (11, 80.81) (13, 79.54) (15, 79.80) (17, 78.09) (19, 79.78)  (21, 79.45) (23, 78.70)  (25, 79.74)(27, 77.85)};
\addplot[
    color=red,
    mark=*,
    ]
    coordinates { (5,  73.17) (7,  80.40) (9,  82.89) (11, 82.49) (13, 83.96) (15, 84.97) (17, 86.07) (19, 85.90) (21, 86.82) (23, 85.52) (25, 85.64) (27, 85.18) };                                                                
\addplot[
    color=green,
    mark=triangle*,
    ]
    coordinates { (5,  78.08) (7,  80.51) (9,  82.97) (11, 85.98) (13, 85.04) (15, 86.78) (17, 86.08) (19, 87.05) (21, 88.19) (23, 87.43) (25, 87.04) (27, 87.35) };                                                                
\end{axis}                                                            
\end{tikzpicture}     

	\caption{
	2D trajectory accuracy Accuracy of 2D trajectory extraction algorithms from left video, right video and both videos for different 2D trajectories: IP, LK and FB}
	\label{fig:IP_LK_FB_Accuracies}                                             
\end{figure*}
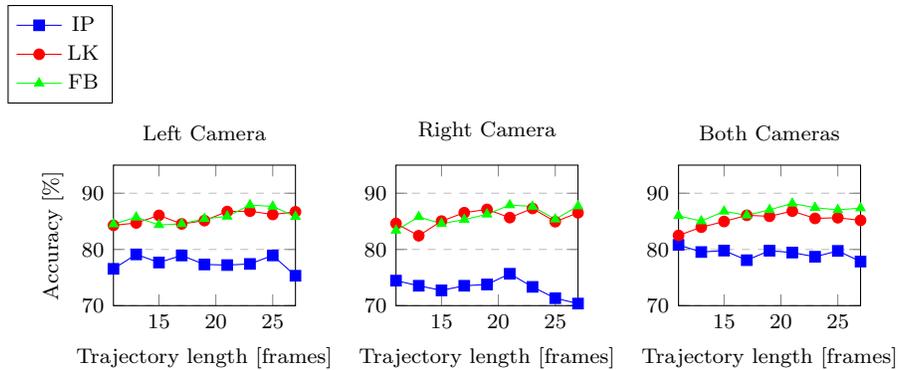

Table~\ref{tab:IP_LK_FB_TrajectoryLength} summarizes the obtained results of our HAR tests using 2D trajectories. Each column represents one of the algorithms proposed in this paper, with ``\emph{both cameras}'' column refers to the simple stacking of left and right descriptors, without any further processing. The descriptors of the trajectories were calculated using the algorithm proposed in~\cite{wang2013dense}.
As it can be seen, FB outperforms the other two in most cases. LK closely follows FB and beats it in some cases. The reason  why FB and LK are yielding similar results is because the selected feature points are the corners, which are easy to follow for both algorithms. Both FB and LK track pixels at subpixel accuracy, yielding smoother trajectories. On the other hand, IP algorithm uses pixel accuracy that degrades its results, as it can be seen on Figure~\ref{fig:IP_LK_FB_Accuracies}.

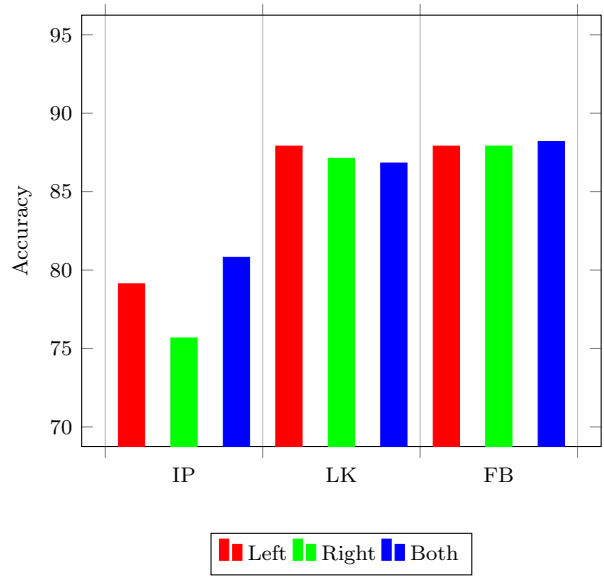
\begin{figure}[!htbp]
	\centering

\begin{tikzpicture}
\begin{axis}[
	width=0.7\linewidth,
	x tick label style={
		/pgf/number format/1000 sep=},
	ylabel={Accuracy},
	xmin=1, xmax=4,
   	ymin=70, ymax=95,
    	xticklabels={IP, LK, FB},
	ytick={70, 75, 80, 85, 90, 95},
	enlargelimits=0.05,
	legend style={at={(0.5,-0.2)},
	anchor=north,legend columns=-1},
	ybar interval=0.5,
]

\addplot[color=red, fill] 
	coordinates {(1, 79.12) (2, 87.90) (3, 87.90) (4,90)};
\addplot [color=green, fill] 
	coordinates {(1, 75.67) (2, 87.12) (3, 87.90) (4,90)};
\addplot [color=blue, fill] 
	coordinates {(1, 80.81) (2, 86.82) (3, 88.19) (4,90)};
\legend{Left,Right,Both}
\end{axis}
\end{tikzpicture}
	\caption{
	The accuracy for left, right and both cameras (best seen in color). The stacking of left and right 2D trajectories barely improve the result.}
	\label{fig::barChartTrajectoriesLeftRightBoth}
\end{figure}

The optimum trajectory length was 21 or 23 for LK and FB. Figure~\ref{fig::barChartTrajectoriesLeftRightBoth} compares the best results obtained from left, right and both cameras. As it can be seen, there is no significant difference between them. In other words, adding up trajectories seen by the left and right cameras did not improve the results. 

\subsection{Effect of Shape Descriptor}

\begin{figure}[!htbp]
\centering
\begin{tikzpicture}
\begin{axis}[
    width = 0.8\linewidth,
    xlabel={Encoding Algorithm},
    ylabel={Accuracy [\%]},
    xmin=1, xmax=7,
    ymin=70, ymax=95,
    xtick={1,2,3,4,5,6,7},
    xticklabels={$D_1$, $D_2$, $D_3$, $D_4$, $D_5$, $D_6$, $D_7$ },
    legend pos=south east,
    ymajorgrids=true,
    grid style=dashed,
]
\addplot[color=red, mark=*] 
	coordinates { (1, 84.50) (2, 85.80) (3, 85.30) (4, 84.84) (5, 84.08) (6, 84.79) (7, 82.10)};
\addplot[color=blue, mark=triangle*] 
	coordinates { (1, 85.72) (2, 87.46) (3, 87.95) (4, 88.42) (5, 85.68) (6, 86.18) (7, 83.15)};
\addplot[color=green, mark=square*] 
	coordinates { (1, 84.37) (2, 86.48) (3, 87.07) (4, 88.55) (5, 88.21) (6, 86.44) (7, 84.96)};
\addplot[color=cyan, mark=x] 
	coordinates { (1, 84.54) (2, 86.94) (3, 89.09) (4, 88.42) (5, 87.11) (6, 86.20) (7, 85.43)};
\legend{Trajectory Length 11,Trajectory Length 13, Trajectory Length 15,Trajectory Length 17}
\end{axis}
\end{tikzpicture}
	
\caption{
The effect of encoding on the performance of Farnback trajectories (best seen in color). The concatenation of higher order derivatives should improve performance, but after 4th derivative the noise level outperforms the benefits.}	
\label{fig::encodingEffectOnPerformance}
\end{figure}
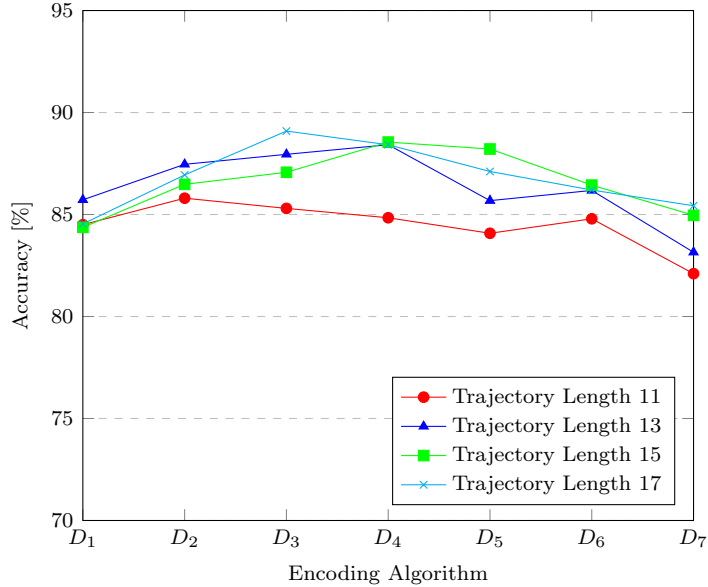

\begin{table*}
  \centering
  \caption{Results of the proposed shape descriptor algorithm on 2D Trajectory (Accuracy \%)}
  \label{tab:TrajectoryEncodingImprovementResults}
  \begin{tabular}
    {cccccccc}
    \toprule
  	& \multicolumn{7}{c}{Trajectory Shape Descriptor} \\
	 	 \cmidrule(lr){2-8}
    	Length  &   $D_1$&  $D_2$ &    $D_3$   &   $D_4$   &  $D_5$ & $D_6$& $D_7$ \\ 
			\cmidrule(lr){1-1}  \cmidrule(lr){2-8}  
 	        11  &    84.50 &    85.80 &    85.30 &   84.84 &  84.08 & 84.79& 82.10 \\
    			13  &    \textbf{85.72} &    \textbf{87.46} &    87.95 &   88.42 &  85.68 & 86.18& 83.15 \\
    			15  &    84.37 &    86.48 &    87.07 &   \textbf{88.55} &  \textbf{88.21} & \textbf{86.44}& 84.96 \\
    			17  &    84.54 &    86.94 &    \textbf{89.09} &   88.42 &  87.11 & 86.20& \textbf{85.43} \\
    \bottomrule
  \end{tabular}
\end{table*}


The proposed trajectory shape descriptor in this paper has improved the classification results. Table~\ref{tab:TrajectoryEncodingImprovementResults} shows the effect of the new algorithm on the accuracy. In the absence of noise, higher order derivatives might provide new information. So, higher order descriptors should produce better results in general. However, in practice the effect of noise and outliers is amplified by the derivatives. Besides, human activities do not have very complex motions. As a consequence, the accuracy has a local maximum bound. Table~\ref{tab:TrajectoryEncodingImprovementResults} and Figure~\ref{fig::encodingEffectOnPerformance} illustrate this effect. As it can be seen, $D_3$ and $D_4$ produced the best accuracy for trajectories, with length 17 and 15, respectively. As expected, higher order derivatives did not improve the performance.

\subsection{Disparity-Augmented Trajectories}
\label{3dExperiments}
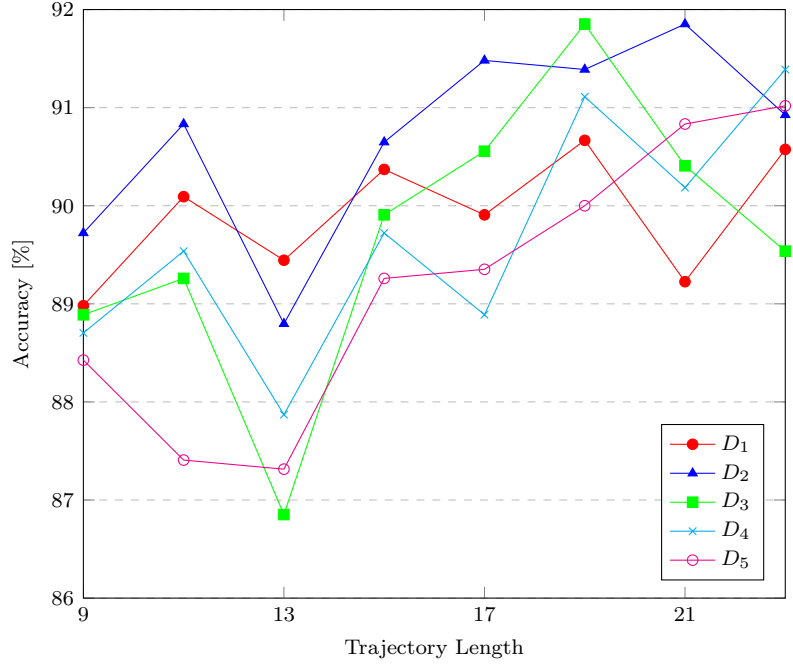
\begin{figure}[!htbp]
	\centering
	\begin{tikzpicture}
		\begin{axis}[
				width = 0.9\linewidth,
				xlabel={Trajectory Length },
				ylabel={Accuracy [\%]},
				xmin=9, xmax=23,
				ymin=86, ymax=92,
				xtick={9, 13, 17, 21},
				legend pos=south east,
				ymajorgrids=true,
				grid style=dashed,
				]
				\addplot[color=red, mark=*]
				coordinates {(9, 88.9815) (11, 90.0926) (13, 89.4444) (15, 90.3704) (17, 89.9074) (19, 90.6667) (21, 89.2256) (23, 90.5741)};
			
				\addplot[color=blue, mark=triangle*]
				coordinates {(9, 89.7222) (11, 90.8333) (13, 88.7963) (15, 90.6481) (17, 91.4815) (19, 91.3889) (21, 91.8519) (23, 90.9259)};
				
				\addplot[color=green, mark=square*]
				coordinates {(9, 88.8889) (11, 89.2593) (13, 86.8519) (15, 89.9074) (17, 90.5556) (19, 91.8519) (21, 90.4074) (23, 89.537)};
		
				\addplot[color=cyan, mark=x]
				coordinates {(9, 88.7037) (11, 89.5370) (13, 87.8704) (15, 89.7222) (17, 88.8889) (19, 91.1111) (21, 90.1852) (23, 91.3889)};
	
				\addplot[color=magenta, mark=o]
				coordinates {(9, 88.4259) (11, 87.4074) (13, 87.3148) (15, 89.2593) (17, 89.3519) (19, 90) (21, 90.8333) (23, 91.0185)};

				\legend{$D_1$, $D_2$, $D_3$, $D_4$, $D_5$}
		\end{axis}
	\end{tikzpicture}

	\caption{
	Disparity-augmented trajectories accuracy. The effect of trajectory length on the accuracy of different encoding algorithms. }
	\label{fig::allTrajectories}
\end{figure}

Table~\ref{tab::disparityResutFinal} summarizes the obtained results for disparity-augmented trajectories. Each row represents a trajectory length while each column represents an encoding algorithm. We have tested trajectory lengths that range between 9 and 27, and encoding up to the fifth degree. As it can be seen, the added disparity information increased the accuracy by around 2\% in all cases. The best obtained result was for trajectory length 19 and encoding degree three. The general trend is that increasing the length of trajectory increases the accuracy of the classification. This trend is more obvious on Figure~\ref{fig::allTrajectories}. The best results (91.85\%) were obtained with $D_2$ and $D_3$ encoding, at trajectory lengths of 21 and 19, respectively.


\begin{table*}
\centering
	\caption{results of disparity-augmented trajectories for human activity recognition}
\label{tab::disparityResutFinal}
\begin{tabular}{cccccc}
	\toprule
	& \multicolumn{5}{c}{Trajectory Shape Descriptor} \\
	  \cmidrule(lr){2-6}
	Length &  $D_1$  &   $D_2$  &  $D_3$   &  $D_4$  & $D_5$    \\
	\cmidrule(lr){1-1}  \cmidrule(lr){2-6}
	9 	          & 88.98\% &   89.72\%  &  88.89\%   &   88.70\%  & 88.43\%  \\
	11 	          & 90.09\% &   90.83\%  &  89.26\%   &   89.54\%  & 87.41\%  \\
	13	          & 89.44\% &   88.80\%  &  86.85\%   &   87.87\%  & 87.31\%  \\
	15	          & 90.37\% &   90.65\%  &  89.91\%   &   89.72\%  & 89.26\%  \\
	17	          & 89.91\% &   91.48\%  &  90.56\%   &   88.89\%  & 89.35\%  \\
	19	          & \textbf{90.67}\% &   91.39\%  &  \textbf{91.85}\%   &   91.11\%  & 90.00\%  \\
	21	          & 89.23\% &   \textbf{91.85}\%  &  90.41\%   &   90.19\%  & 90.83\%  \\
	23	          & 90.57\% &   90.93\%  &  89.54\%   &   \textbf{91.39}\%  & \textbf{91.02}\%  \\
	\bottomrule
\end{tabular}
\end{table*}

\begin{figure*}[!htbp]
	\centering
	\includegraphics[width=1.0\linewidth]{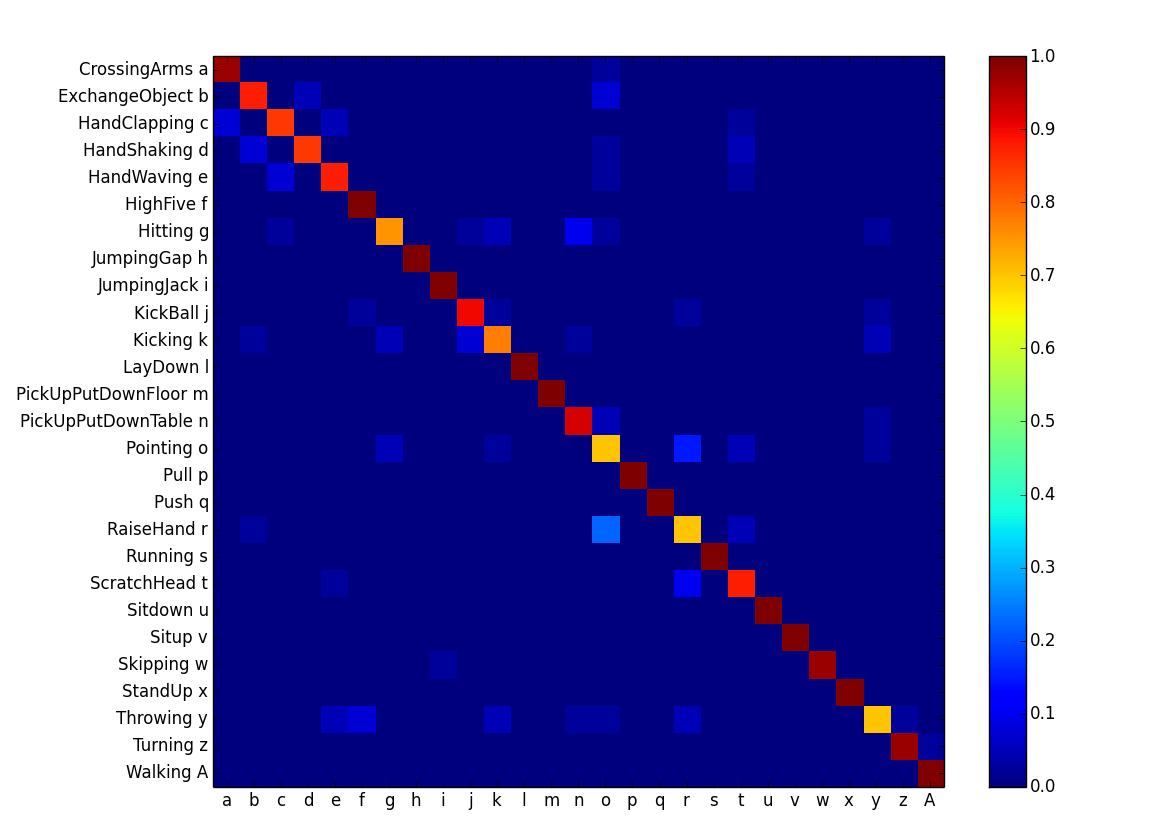}
	\caption{
	The confusion matrix. Sample confusion matrix for 27 classes (indexed a to z and A). Each row represents the actual class, and each column is the predicted class.}
	\label{fig::confusionMatrix}
\end{figure*}

Figure~\ref{fig::confusionMatrix} illustrates the confusion matrix of a sample test. Each row represents an actual class and each column represents a predicted class. The number of correct classifications is normalized between zero and one. It is also worth noting that the classes in our dataset are balanced, i.e., the number of samples for each activity classes is the same for all classes. The misclassified instances from the matrix give interesting information about the behavior of trajectories for HAR. For example, the most confused classes in this figure are ``pointing'' and ``raise hand''. The fact that for pointing to something, one should raise his/her hand shows that trajectories are capable of finding this similarity, but they are unable to distinguish between them in some cases. Another example is the classes ``kicking a fixed object'' and ``kicking the ball''. These classes have very similar motions and hence, they are expected to be confused by any motion-based HAR method. 

From another viewpoint, it can be assumed that human activities have no precise definitions. In particular, many human activities do have some overlaps. For example, raising hand to point to something or waving. So, it is evident that there is a conceptual overlap over the definition of these classes and it is not easy to separate them conceptually.

\subsection{Time measurements}
Our comparisons have shown that our disparity-augmented trajectory (DAT) method is faster than trajectory aligned methods (see Table~\ref{tab::timing}). 
To test the time performance, five different random samples from the dataset were selected. Each reported time is the average of ten different runs. As it can be seen, the speed up gains vary between 1.5 to 4.5 times faster, depending on the selected activity samples. Overall, our proposed method is more than twice faster than the methods based on dense trajectories, as we are using sparse features.

\begin{table*}
\centering
\caption{Time improvements obtained by using our method over trajectory aligned descriptors for different random samples, taken from our dataset.}
\label{tab::timing}
\begin{tabular}{lcccccc}
\toprule
Method	 &   S1   &  S2 & S3 & S4 & S5 & sum \\
\midrule
DAT(our method)		 &   19.67s   & 32.13s  &  7.33s & 42.70s & 13.09s & 114.92s \\
Trajectory Aligned    &  46.89s & 51.95s & 33.68s & 66.17s & 41.31s & 240.00s \\
\midrule
Speed up &  238\%	  &  161\%  &  459\%  & 155\%  & 311\% & 209\%  \\
\bottomrule
\end{tabular}
\end{table*}


\subsection{Comparison}
Table~\ref{tab::comparison} shows the performance of our proposed method compared to the state of the art. The closest works to DAT are 2D dense trajectories~\cite{wang2013action} and 3D trajectories~\cite{koperski20143d}.
We applied the algorithm proposed in~\cite{wang2011action} on sparse feature points and the result reported as sparse trajectories in Table~\ref{tab::comparison}. As it can be seen, our proposed method was able to outperform both dense and sparse trajectory methods, with a good margin. Moreover, the proposed method produced better result compared to HOG, and competitive results to HOF and MBH. It should be noted that HOG, HOF and MBH need more computation time in comparison with our disparity-augmented trajectories.

Koperski et al.\cite{koperski20143d} used depth information to create 3D trajectories. They ran their tests on \emph{MSR DailyActivity 3D} dataset, which has similar setting as our dataset, but recorded with an RGB-Depth camera. Because we are using disparity and not depth, it is not possible to run their algorithm on our dataset, and we cannot run our method on their dataset as well. Just for comparison, we reported their results in Table~\ref{tab::koperski}. As it can be seen, they could not improve the performance of 2D trajectories by using 3D data only. They only improved the performance by combining 2D and 3D data. This confirms again the effectiveness of our proposed trajectory shape descriptor. 

\begin{table}
\centering
\caption{Comparison of our method against other state-of-the-art methods}
\label{tab::comparison}
\begin{tabular}{rlc}
\toprule
& Method					& Accuracy  \\
\midrule
\multirow{3}{*}{Trajectory based}
& Sparse Trajectories   &   87.80\% \\
& Dense Trajectory      &	88.74\% \\
& \textbf{DAT} (Ours)	&	91.85\%	\\
\midrule
\multirow{3}{*}{Trajectory aligned}
& HOG                   &	89.54\%	\\
& HOF					&	92.72\%	\\
& MBH					&	92.22\%	\\
\bottomrule

\end{tabular}
\end{table}

\begin{table}
\centering
\caption{3D trajectories proposed by Koperski et al.~\cite{koperski20143d}.}
\label{tab::koperski}
\begin{tabular}{lc}
\toprule
Method			& Accuracy  \\
\midrule
2D TSD  		&  78\% \\
3D TSD   		&  74\% \\
2D TSD+3D TSD	&  85\%	\\
\bottomrule
\end{tabular}
\end{table}


\begin{table}
\centering
\caption{Comparisons of the obtained results on Hollywood 3D dataset. The first 5 rows are our DAT results with different trajectory lengths.}
\label{tab::hollywood3D}
\begin{tabular}{lc}
\toprule
length	& Accuracy  \\
\midrule
7		&  21.50\%  \\
9   	&  22.73\%  \\
11		&  \textbf{23.78\%}	\\
13		&  20.79\%	\\
15		&  19.80\%	\\
\midrule
\cite{wang2013dense} & 20.8\%\\
\cite{hadfield2017hollywood}    &  21.8\% \\
\bottomrule
\end{tabular}
\end{table}

We have also tested our method on Hollywood 3D dataset, but have used their disparities only. Table~\ref{tab::hollywood3D} shows our obtained results and their comparisons with the other relevant methods. Note that our proposed method was not designed to cope specifically with random camera motions and/or rotations, as they can degrade the trajectory extraction drastically. 
As it can be seen, our method still yields superior results compared to the trajectory aligned descriptors proposed in \cite{wang2013dense} and reported in \cite{hadfield2017hollywood}. Our method also outperforms the method proposed by \cite{hadfield2017hollywood} in terms of accuracy.


\section{Conclusion}
\label{sec::conclusion}
We have presented and compared three popular trajectory-based human action recognition methods. We have also enhanced the conventional trajectory encoding algorithms by considering higher order derivatives of individual trajectories. 
Furthermore, we have proposed a new method based on disparity-augmented trajectories for video content analysis. Because disparities carry the scene's three-dimensional clues, we anticipated an improvement in the HAR performance.
In particular, we have fused the disparity information with motion-based features. 
We have obtained improved results on HAR, when compared to traditional trajectory-based methods, at a lower computational cost.

Furthermore, we have demonstrated that trajectories are useful for video content analysis in general, and for human activity recognition in particular. The proposed shape encoding algorithm has improved the accuracy of activity recognition by about 1.5\%. The disparity information added to trajectories has also enhanced the results by another 2.5\%.

We have also discussed some limitations associated with trajectory-based activity recognition. Activities that are similar, from the movement point of view, might be confused. We believe that some actions are conceptually overlapping and are hard to be distinguished, when using the human movement information only.


\bibliography{HAR}

Author Biographies:

\textbf{Pejman Habashi} received his M.sc. in artificial intelligence from Sharif University of Technology, Tehran, Iran in 2008. He received his Ph.D. in computer scient from University of Windsor, Ontario, Canada in 2018. Previously he worked as a research assistant at the University of Windsor.

\textbf{Boubakeur Boufama} received his M.Sc. degree in Computer Science from the University of Nancy I, France, in 1991, and his Ph.D., also in Computer Science, from the Grenoble Institute of Technology, France, in 1994. He is currently a professor with the School of Computer Science at the University of Windsor, Ontario, Canada.

\textbf{Imran Shafiq Ahmad} received M.Sc. in Applied Physics from the University of Karachi-Pakistan in1986 and M.S. and Ph.D.  in Computer Science from the Central Michigan University and Wayne State University-USA in 1992 and 1997 respectively.  He is currently an Associate Professor in the School of Computer Science at the University of Windsor, Ontario-Canada.

\end{document}